\documentclass[letterpaper, 10 pt, journal, twoside]{ieeetran}
\usepackage{amsmath,amsfonts}
\usepackage{array}
\usepackage[caption=false,font=normalsize,labelfont=sf,textfont=sf]{subfig}
\usepackage{textcomp}
\usepackage{color}
\usepackage{stfloats}
\usepackage{url}
\usepackage{verbatim}
\usepackage{amsmath}
\usepackage{graphicx}
\usepackage{booktabs}
\usepackage{siunitx}
\usepackage{booktabs}
\usepackage{multirow}
\usepackage{bm}
\usepackage{cite}
\usepackage{colortbl}
\usepackage{hyperref}
\usepackage{xcolor, soul}
\sethlcolor{red}
\usepackage{tabularx}
\usepackage{colortbl} % 关键，支持行着色
\usepackage{xcolor}    % 颜色支持
\usepackage{algorithm}        % 算法浮动体
\usepackage{algpseudocode}    % 伪代码样式
\usepackage{setspace}
\usepackage{array} 
\usepackage{booktabs} % 用于 \Xhline

\definecolor{LightGray}{gray}{0.9} % 定义浅灰色

\hypersetup{hypertex=true,
colorlinks=true,
linkcolor=blue,
anchorcolor=blue,
citecolor=blue}

\def\BibTeX{{\rm B\kern-.05em{\sc i\kern-.025em b}\kern-.08em
    T\kern-.1667em\lower.7ex\hbox{E}\kern-.125emX}}
\usepackage{balance}

\begin{document}

\title{Riemannian Optimization-based Whole-body Obstacle Avoidance for Mobile Continuum Robots in Cluttered Environments}

\author
{Duo Tang$^1$, Ping Deng$^1$, Sophie Armanini$^2$, Yajing Shen$^{3\dagger}$,\\ Toshio Fukuda$^4$, Cosimo Della Santina$^5$, Peng Lu$^{1\dagger}$\\
\normalsize{$^1$Department of Mechanical Engineering, University of Hong Kong, Hong Kong, China}\\
\normalsize{$^2$Department of Aeronautics, Imperial College London, UK}\\
\normalsize{$^3$Department of Electronic and Computer Engineering, }\\
\normalsize{Hong Kong University of Science and Technology, Hong Kong, China}\\
\normalsize{$^4$Nagoya University, Meijo University, Japan and Beijing Institute of Technology, China}\\
\normalsize{$^5$Department of Cognitive Robotics, Delft University of Technology, Delft, The Netherlands}\\
%\normalsize{$\dagger$These authors contributed equally to this work.}\\
\normalsize{$^\ast$Corresponding author. Email: tangduo777@connect.hku.hk, lupeng@hku.hk}
}

% \author
% {Anonymous authors}

% Paper headers
\markboth{}  
{\MakeLowercase{\textit{et al.}}: Collision-Free Motion Planning and Control for Mobile Continuum Robots} 
% Use only for the final RAL version

\maketitle
\begin{abstract}

In nature, snakes, elephant trunks, and octopus arms demonstrate the ability to traverse cluttered and confined environments with remarkable agility, relying on tight coupling between global behavioral intent and local morphological adaptation. This characteristic provides important inspiration for motion planning and control of continuum robots. However, existing approaches for mobile-base continuum robots typically emphasize either high-level path planning or local motion control in isolation, making it difficult to simultaneously ensure dynamic feasibility, kinematic consistency, and operational safety in narrow passages and geometrically complex environments. To address this challenge, we propose a unified planning and control framework for mobile-base continuum robots that enables coordinated long-horizon planning and real-time execution. The framework integrates task-level trajectory reasoning with geometry-aware local motion adaptation, achieving consistent global-to-local decision making under kinematic and environmental constraints. The global layer employs an A*-guided minimum-jerk planning scheme to generate dynamically feasible long-horizon reference trajectories, while the local layer ensures real-time kinematic feasibility and collision avoidance through a geometry-consistent inverse kinematics solver and a Riemannian manifold-based optimization strategy. Extensive validation is conducted in both simulation and real-world systems. In simulation, pipe traversal tasks are designed and evaluated against multiple baseline methods. Real-world experiments include navigation in obstacle-dense environments and manipulation in confined spaces. The results demonstrate that the proposed method improves motion consistency, feasibility, and environmental adaptability for mobile continuum robots operating in constrained environments, providing an effective technical pathway toward real-world deployment.

\end{abstract}


\section{Introduction}

Biological systems with compliant bodies navigate cluttered environments by exploiting their intrinsic kinematic redundancy and distributed sensing. Snakes, octopus arms, and elephant trunks, despite their distinct morphologies, achieve robust locomotion and interaction through continuous body deformation and physical interaction with their surroundings \cite{Jayne1986, Astley2015, Gutnick2020, Dagenais2021, Schulz2023}. Inspired by these biological structures, continuum robots replace discrete rigid-link architectures with bodies that bend and twist continuously \cite{Webster2010, Sakuhara2020, peng2025dexterous, Wright2012}. Their inherent compliance and high kinematic redundancy enable access to environments that are difficult for rigid manipulators to reach, including tortuous anatomical pathways in surgical settings \cite{Shi2017, Hwang2019}, collapsed structures in search-and-rescue scenarios \cite{Yamauchi2022, Wang2021}, and confined spaces for inspection and maintenance \cite{Zheng2013, Xu2018}. However, translating these morphological advantages into autonomous operation in unstructured environments remains challenging. The continuous deformation of the robot body introduces highly nonlinear kinematics and dynamics, making accurate modeling, inverse kinematics, and collision-aware motion generation considerably more difficult than for rigid-link manipulators \cite{DellaSantina2020IJRR}. 

In practical environments, a continuum manipulator cannot always rely on a fixed base to provide a suitable workspace. Mounting the manipulator on a mobile platform can substantially enlarge its reachable workspace and enable navigation and manipulation within large, cluttered, and spatially constrained environments. However, mobility also fundamentally changes the planning problem. The motion of the mobile base and the deformation of the continuum arm are strongly coupled: base motion changes the reachable workspace of the end-effector, while arm configuration determines collision constraints and feasible base poses. Consequently, autonomous operation requires planning over a high-dimensional configuration space while simultaneously maintaining geometric consistency of the deformable arm, avoiding obstacles, and satisfying the computational constraints of real-time execution.

Existing research has made substantial progress in continuum robot actuation, kinematic modeling, and motion planning \cite{Yuan2018, Lee2016, Rone2014, Uppalapati2020, Webster2010, Neppalli2009, DellaSantina2020RAL, Ma2023, Chen2023, Tian2021}. However, most studies focus on individual subsystems or operate under restrictive assumptions, with limited consideration of the coupling among mobile-base motion, continuum deformation, and real-time collision avoidance. In particular, coordinated base–arm planning for mobile continuum robots remains relatively underexplored \cite{Boutchouang2022}.

This gap is particularly important for real-world deployment. Independently planning the mobile base and continuum arm can produce globally inconsistent motions, while jointly optimizing their high-dimensional coupled configuration is computationally expensive. We therefore propose a hierarchical motion planning framework that coordinates long-horizon global planning with real-time local geometric refinement. The local layer incorporates a curvature-based inverse kinematics solver that enforces segment-level geometric consistency and a Riemannian orientation optimizer that generates smooth, singularity-free trajectories. Integrated with a compact mobile continuum robot platform, the proposed system enables coordinated base-arm motion with real-time collision avoidance in cluttered environments.

The main contributions of this work are summarized as follows:

\begin{enumerate}
\item An integrated platform for mobile continuum robots is developed, providing a compact and lightweight physical system with integrated mechanical structure, actuation, and onboard electronics for autonomous navigation and manipulation.

\item A hierarchical motion planning framework is proposed to address the high-dimensional configuration space and coupled base–arm motion of mobile continuum robots. The framework separates global trajectory generation from local geometric refinement while coordinating mobile base motion with continuum arm deformation, enabling real-time planning in cluttered environments.

\item Two core algorithmic components are developed for local planning. The first is a curvature-based inverse kinematics solver that reformulates FABRIK to enforce segment-level geometric consistency. The second is a Riemannian orientation optimizer that generates smooth and singularity-free orientation trajectories on the orientation manifold. 

\item Comprehensive simulation and real-world experiments validate the proposed system under cluttered and constrained conditions. Comparisons with baseline methods demonstrate improvements in computational efficiency and trajectory quality, while physical experiments further demonstrate robust coordinated navigation and collision avoidance on the mobile continuum robot.
\end{enumerate}

\section{Related Works}
\label{sec:related_works}
Autonomous planning for mobile continuum robots requires both efficient inverse kinematics and collision-aware motion planning, owing to their high redundancy and nonlinear deformation \cite{Russo2023}. Existing studies have therefore primarily addressed these two aspects separately, with efforts focusing on real-time inverse kinematics \cite{Ataka20162} and obstacle avoidance \cite{Yang2024}. 

\subsection{Inverse Kinematics Solver}

Unlike rigid-link robots, continuum robots achieve motion through continuous bending, making their kinematic modeling and control more challenging \cite{Mahl2014}. Various kinematic models have therefore been developed to improve the accuracy and computational efficiency of inverse kinematics \cite{Tang2019, Zhang2019}.

The constant curvature (CC) assumption is widely adopted, approximating each segment as a circular arc to simplify kinematic formulation and reduce computational cost \cite{Webster2010, Neppalli2009}. However, deviations caused by external forces, material properties, and environmental interactions can reduce its accuracy. More advanced models incorporating physical constraints and external loads provide improved accuracy \cite{Jones2006KinematicsFM}, but at increased computational cost, which can limit their use in real-time applications.

Based on these models, various inverse kinematics methods have been developed. FABRIK (Forward and Backward Reaching Inverse Kinematics) has been widely adopted for its computational efficiency and singularity avoidance \cite{Kolpashchikov2018, Zhang2018, aristidou2011fabrik}, with extensions such as FABRIKx addressing variable-curvature continuum robots \cite{kolpashchikov2022}. However, these methods primarily focus on end-effector positioning and do not explicitly account for whole-body collision constraints, limiting their applicability in cluttered environments.

Closed-form analytical approaches provide fast and deterministic solutions \cite{Neppalli2009}, while geometric methods exploit structural constraints to simplify inverse kinematics \cite{Neppalli2008AGA}. Their reliance on simplified kinematic assumptions, however, can limit performance when significant deformation or complex environmental interactions are present.

Learning-based approaches have also been explored to capture nonlinear deformation and varying operating conditions \cite{Thuruthel2016LearningGI}. Although they can reduce reliance on explicit kinematic models, their generalization to unseen configurations and environments remains a challenge for real-time autonomous operation.

\subsection{Obstacle Avoidance Methods}

Obstacle avoidance for continuum robots is challenging due to their high redundancy and nonlinear deformation \cite{Carriker1991}. Optimization-based methods are widely used to incorporate environmental and kinematic constraints \cite{Gao2020}. MPC-based approaches further enable receding-horizon trajectory optimization and have been extended to dynamic obstacles, soft constraints, and hierarchical planning \cite{Hachen2025, Qin2025, Qin2024}. Interior-point optimization has also been applied to generate collision-free trajectories for hybrid continuum robots \cite{Ma2023}. However, the computational cost of these methods can increase substantially with the dimensionality of the configuration space.

Sampling-based methods improve exploration of high-dimensional configuration spaces through guided sampling and constraint integration \cite{Chen2023, li2024, Luo2024}. Nevertheless, their probabilistic nature and reduced sampling efficiency in narrow passages can limit their use in densely cluttered environments requiring fast and deterministic planning.

Artificial potential field (APF) methods provide computationally efficient reactive avoidance and have been combined with search algorithms and null-space control for dynamic and surgical applications \cite{Tian2021, Ding2024, Li2023}. However, susceptibility to local minima can reduce their reliability in complex cluttered environments.

Compliance- and deformation-based approaches exploit the flexibility of continuum robots for safe interaction, including contact-force control, haptic feedback, and learning-based shape estimation \cite{Xu2019, Xu2024, Mohammadreza2025}. These methods can improve reactive adaptation but generally provide limited guarantees of collision-free motion and may depend on accurate deformation estimation.

    \begin{figure*}[t]
    \centering    \includegraphics[width=0.9\linewidth]{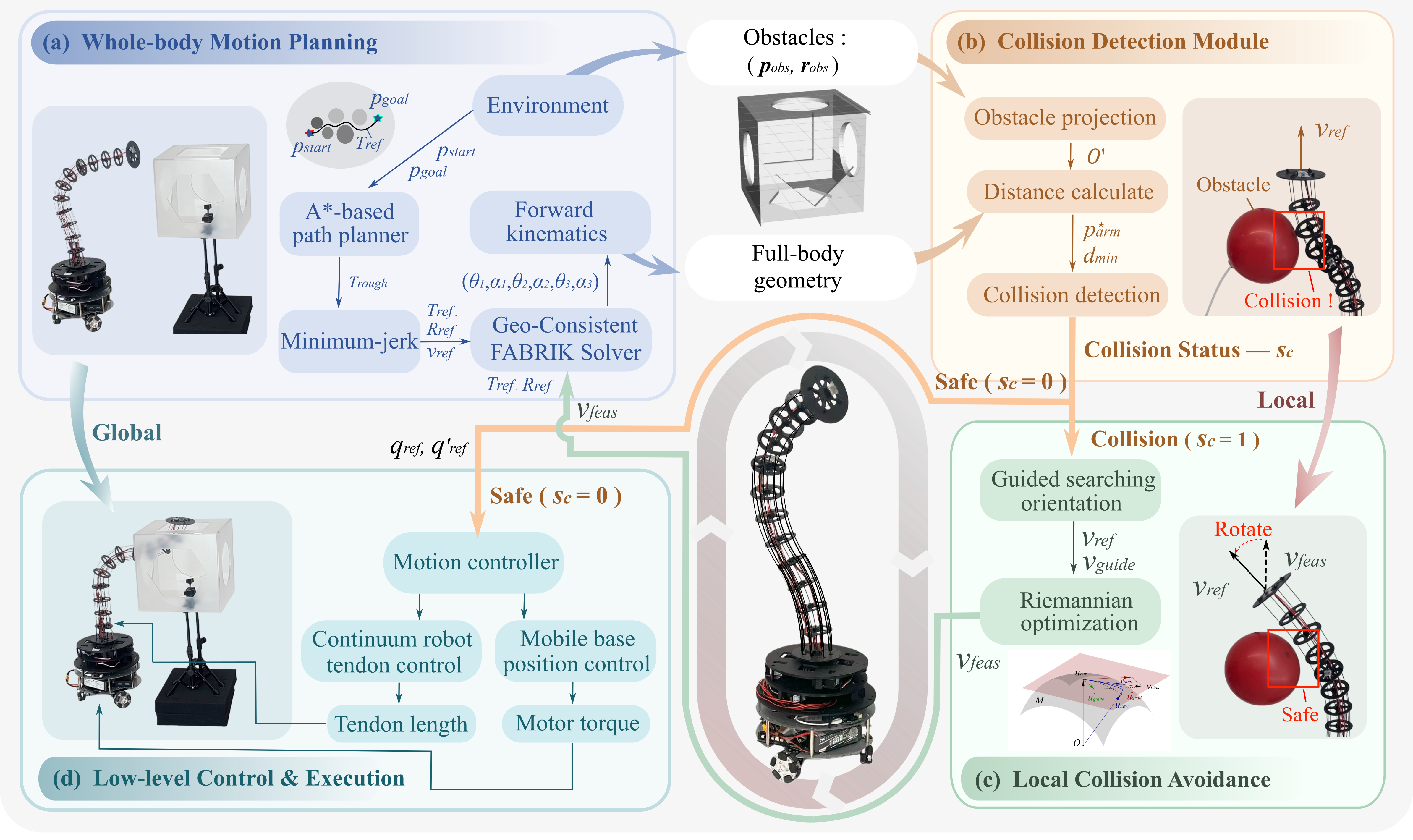}
    \caption{Overall framework of the proposed motion planning and execution pipeline for a mobile-base continuum robot in cluttered environments. (a) Whole-body motion planning module: Generates globally feasible whole-body trajectories toward the task goal. (b) Collision detection module: Continuously detects potential collisions between the robot and the environment. (c) Local collision avoidance module: Searches the orientation space and modifies the motion to avoid collisions with obstacles. (d) Low-level control and execution module: Executes control commands to track the planned trajectories.}
    \label{Framework}
\end{figure*}

\section{Overall Framework}
\label{sec:framework}

The overall framework of the proposed motion planning and execution pipeline for a mobile continuum robot in cluttered environments is illustrated in Fig.~\ref{Framework}. 

\subsection{Whole-body Motion Planning Module}

The whole-body motion planning module forms the top layer of the framework. It first extracts the start and goal states from the environment and task objectives. A preliminary collision-free path $\bm{T_{\text{rough}}}$ is generated using an A* planner\cite{tang2021geometric}, which is then refined via minimum-jerk smoothing to produce a continuous reference trajectory $T_{\text{ref}}$ with end-effector orientation $\bm{v_{\text{ref}}}$ and rotation matrix $\bm{R_{\text{ref}}}$\cite{12075}. This task-space trajectory is solved into a sequence of configuration variables $q$ using an iterative geometry-consistent FABRIK-based inverse kinematics solver, where $q$ is defined under the piecewise constant curvature representation of the continuum manipulator and represents a geometric shape parameterization used for planning rather than an actuation-level control variable. For body-level collision detection and local avoidance, forward kinematics reconstructs the full-body configuration of the manipulator in the environment, which cannot be inferred from the end-effector pose alone. 

\subsection{Collision Detection Module}

The collision detection module ensures safe manipulator motion. Obstacle information is obtained from the known environment and approximated using a set of bounding spheres, where each obstacle is represented by multiple spheres defined by their center positions $\bm{p_{\text{obs}}}$ and radii $r_{\text{obs}}$. The collision status is determined as $s_c = 0$ or $1$, where $s_c = 0$ indicates a collision, triggering local avoidance, and $s_c = 1$ indicates no collision, allowing execution to proceed.

\subsection{Local Collision Avoidance Module}
The local avoidance module computes a guided direction $\bm{v_{\text{guide}}}$ from the obstacle position $\bm{p_{\text{obs}}}$ and manipulator configuration $G$ as the initial step for collision-free planning. Guided by $\bm{v_{\text{guide}}}$, a Riemannian optimization search starting from $\bm{v_{\text{ref}}}$ produces a feasible execution orientation $\bm{v_{\text{feas}}}$, which is fed back to the inverse kinematics and forward kinematics modules to check for collisions along the original target positions. If a collision remains, the search iterates; once $\bm{v_{\text{feas}}}$ satisfies collision-free conditions, the trajectory proceeds to execution.

\subsection{Low-level Control and Execution Module}
Once the manipulator can move along the target position with orientation $(\bm{v_{\text{ref}}}$ or $\bm{v_{\text{feas}}})$ without collision, the low-level module executes joint-space commands for tendon lengths and mobile base motion. The continuum robot tendon controller generates tendon length signals, and the mobile base position controller produces motor torques. This module ensures accurate tracking and stable and safe execution of the planned collision-free motion.

\section{Collision-free Whole-body Motion Planning Algorithm}
\label{sec:motion_planning}

In this section, we propose a novel collision avoidance motion planning method for mobile continuum robots. This approach first performs precise and real-time collision detection to identify potential risks, then constructs an artificial potential field to generate guided searching orientation, and finally applies Riemannian manifold-based optimization on the task space to improve search efficiency and ensure smooth motion.

\subsection{Problem Formulation}  
\label{subsec:problem_formulation}
The motion planning and control of the continuum robot in cluttered environments can be formulated as a nonlinear optimal control problem, where both collision-free motion and goal-reaching accuracy must be simultaneously guaranteed.  
We first define the configuration space of a three-section continuum robot as 
$\bm{\Theta} = [\theta_1, \alpha_1, \theta_2, \alpha_2, \theta_3, \alpha_3]^T$.

The joint-space optimization problem for motion planning can be expressed as:  

\begin{equation}
\label{eq:joint_space_formulation_refined}
\left\{
\begin{aligned}
& \text{find}    && \bm{\Theta} \in \mathbb{T}^6 \\[2pt]
& \text{minimize}    && \bm{\mathcal{F}}(\bm{\Theta}) = 
    \begin{bmatrix} 
    \left( 1 - \bm{u}(\bm{\Theta})^T \bm{u}_\mathrm{des} \right)^2 \\
    \|\bm{p}_{\mathrm{end}}(\bm{\Theta}) - \bm{p}_\mathrm{target}\|^2 
    \end{bmatrix} \\[4pt]
& \text{subject to} &&
\begin{cases}
\mathrm{dist}(\bm{\xi}(\bm{\Theta}), \bm{P}_i) \geq r_i, \quad \forall i \\
\arccos\!\left( \bm{u}(\bm{\Theta}_{k+1})^T \bm{u}(\bm{\Theta}_k) \right) \leq \Delta\bm{\Theta}_{\max} \\
\|\bm{p}_{\mathrm{end}}(\bm{\Theta}) - \bm{p}_\mathrm{bound}\| \leq \Delta \bm{p}_{\max} 
\end{cases}
\end{aligned}
\right.
\end{equation}  

The objective function $\bm{\mathcal{F}}(\bm{\Theta})$ measures both end-effector orientation and position errors. Here, $\bm{u}(\bm{\Theta}) \in \mathbb{R}^3$ is the unit direction of the end-effector, $\bm{u}_{\mathrm{des}}$ is the desired orientation, $\bm{p}_{\mathrm{end}}(\bm{\Theta})$ is the current end-effector position, and $\bm{p}_{\mathrm{target}}$ is the target position.

The optimization is subject to several constraints to ensure safe and feasible motions. Collision avoidance requires the robot backbone $\bm{\xi}(\bm{\Theta})$ to stay at least $r_i$, the safety radius of obstacle $i$, away from each obstacle $\bm{P}i$. Smoothness is enforced by limiting changes in end-effector orientation between consecutive steps using $\Delta \bm{\Theta}{\max}$, the maximum allowable orientation change. Workspace feasibility restricts the end-effector position within a bounded region around $\bm{p}_{\mathrm{bound}}$ with tolerance $\Delta \bm{p}_{\max}$, the maximum allowable position deviation.

To improve computational efficiency and avoid the high-dimensional complexity of joint-space optimization, the motion planning problem is reformulated in the task space. Since each desired task pose $(\bm{v}_\mathrm{des}, \bm{p}_\mathrm{target})$ can be mapped to feasible joint configurations via inverse kinematics, it is sufficient to optimize directly over the end-effector orientation $\bm{v} \in \mathbb{S}^2$, with the spherical manifold $\mathbb{S}^2$ providing a compact and continuous representation that naturally avoids singularities.

\begin{equation}
\label{eq:task_optim_refined}
\left\{
\begin{aligned}
& \text{find}    && \bm{v} \in \mathbb{S}^2 \\[2pt]
& \text{minimize} && \mathcal{J}_\mathrm{total}(\bm{v}) \\[4pt]
& \text{subject to} &&
\begin{cases}
d_\mathrm{sdf}(\bm{v}, \bm{O}_i) \geq r_i, \quad \forall i \\
\angle(\bm{v}_{k+1}, \bm{v}_k) \leq \phi_\mathrm{max} \\
\|\bm{p}_\mathrm{ik}(\bm{v}) - \bm{p}_\mathrm{safe}\| \leq \delta_\mathrm{max}
\end{cases}
\end{aligned}
\right.
\end{equation}  

The scalarized objective function
\begin{equation}
\mathcal{J}_\mathrm{total}(\bm{v}) = w_\mathrm{dir} \left(1 - \bm{v}^T \bm{v}_\mathrm{des}\right)^2 + w_\mathrm{pos} \|\bm{p}_\mathrm{ik}(\bm{v}) - \bm{p}_\mathrm{target}\|^2,
\end{equation}
balances orientation alignment and position reaching, where $w_\mathrm{dir}$ and $w_\mathrm{pos}$ are positive weights summing to one, and $\bm{p}_\mathrm{ik}(\bm{v})$ denotes the end-effector position obtained from inverse kinematics given orientation $\bm{v}$. This formulation allows the planner to operate directly on task-relevant variables while implicitly ensuring feasible joint configurations.

\begin{figure}[t]
    \centering
    \includegraphics[width=\linewidth]{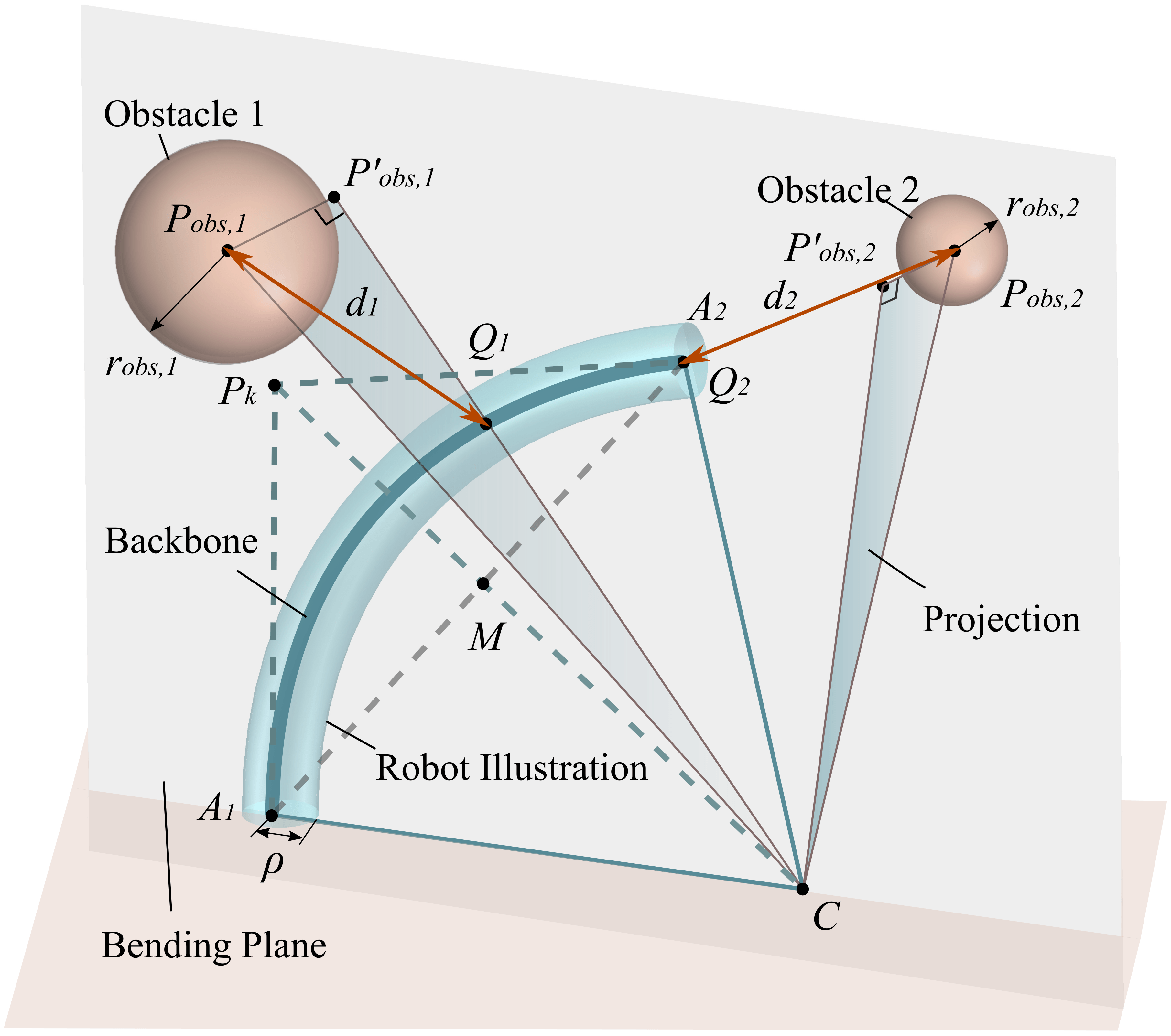}
    \caption{Illustration of the analytical collision checking scheme for continuum robots, showing continuous arc modeling, obstacle projection, and safety boundary evaluation. The method overcomes limitations of discrete sampling approaches through geometric parameterization.}
    \label{fig:collicheck}
\end{figure}

In the task-space problem, $d_\mathrm{sdf}(\bm{v}, \bm{O}_i)$ represents the minimum distance between the current robot configuration and obstacle position $\bm{O}_i$, computed using the collision detection method described in the next section, and $r_i$ denotes the minimum allowed clearance, which is the safety radius of
obstacle $i$. $\bm{v}_k$ and $\bm{v}_{k+1}$ are consecutive end-effector orientations with maximum allowed change $\phi_\mathrm{max}$, and $\bm{p}_\mathrm{ik}(\bm{v})$ is the end-effector position obtained via kinematic mapping, which is constrained within $\delta_\mathrm{max}$ of a reference point $\bm{p}_\mathrm{safe}$ to ensure that the resulting configuration remains within the feasible workspace and avoids generating physically infeasible robot poses.

To solve it efficiently, we adopt an optimization strategy on the orientation manifold. At each step, a feasible guidance orientation is determined through real-time collision detection. Subsequently, the orientation is refined via optimization on the manifold. To ensure the continuity of orientation, each step is initialized with the solution from the previous one. This incremental, manifold-based approach reduces computational complexity, improves convergence robustness.

\subsection{Collision Detection}
\label{subsec:collision_detection}

Accurate and rapid collision detection constitutes a fundamental prerequisite for enabling effective obstacle avoidance of continuum robots. This section presents a geometrically rigorous collision detection method for continuum robots, which overcomes the limitations of discrete sampling approaches through analytical arc parameterization.

The manipulator is decomposed into multiple arc segments under the piecewise constant curvature assumption, as shown in the geometric representation of Fig.~\ref{fig:collicheck}. Each arc segment is defined by the endpoint position vectors $\bm{A}_1, \bm{A}_2 \in \mathbb{R}^3$ and an interpolation point with position vector $\bm{P}_k \in \mathbb{R}^3$, all expressed in a common Cartesian reference frame attached to the robot base. The arc geometry is characterized through parametric derivation. The arc center position $\bm{C}$ is computed as
\begin{equation}
  \bm{C}
  =
  \bm{P}_k
  +
  \frac{\sqrt{\|\bm{P}_k - \bm{A}_1\|^2 + R^2}}{\|\bm{M} - \bm{P}_k\|}
  \left(\bm{M} - \bm{P}_k\right),
\end{equation}
where $\bm{M} = (\bm{A}_1 + \bm{A}_2)/2$ represents the midpoint position between the endpoints and $R$ denotes the curvature radius. (The vector from a point $\bm{X}$ to a point $\bm{Y}$ is represented by the difference of their position vectors, i.e., $\bm{Y}-\bm{X}$.) This formulation preserves the manipulator's continuous geometric properties, eliminating discretization artifacts inherent in sampling-based approximations.

For obstacle interaction analysis, the three-dimensional collision detection problem is reduced to planar geometry through orthogonal projection, corresponding to the projection scheme depicted in Fig.~\ref{fig:collicheck}. The arc plane normal vector is derived using the cross product
\begin{equation}
  \bm{N}
  =
  \frac{ (\bm{A}_1 - \bm{C}) \times (\bm{A}_2 - \bm{C}) }
  { \| (\bm{A}_1 - \bm{C}) \times (\bm{A}_2 - \bm{C}) \| }.
\end{equation}

Specifically, let $\bm{P}_{obs}$ be the obstacle center. We first project it orthogonally onto the bending plane with center $\bm{C}$ and unit normal $\bm{N}$:
\begin{equation}
\bm{P}_{obs}'
=
\bm{P}_{obs}
-
\bigl[(\bm{P}_{obs} - \bm{C}) \cdot \bm{N}\bigr]\bm{N}.
\end{equation}

The closest point $\bm{Q}$ on the arc is obtained as the analytical solution to a constrained point-to-arc distance problem, following the geometric construction illustrated in Fig.~\ref{fig:collicheck}. Specifically, the unconstrained closest point on the supporting circle is first derived in closed form by radial projection within the arc plane:
\begin{equation}
\bm{Q}_{c}
=
\bm{C}
+
R\,\frac{\bm{P}_{obs}' - \bm{C}}{\|\bm{P}_{obs}' - \bm{C}\|}.
\end{equation}
This point, corresponding to $\bm{Q}_c$ in our illustration (as $Q_1$ in Fig.~\ref{fig:collicheck}), represents the unique minimizer of the Euclidean distance to the infinite circular support of the arc.

To enforce the finite arc segment constraint, the admissible angular interval is defined by the endpoint direction vectors
\begin{equation}
\hat{\bm{u}}_1
=
\frac{\bm{A}_1 - \bm{C}}{\|\bm{A}_1 - \bm{C}\|},\quad
\hat{\bm{u}}_2
=
\frac{\bm{A}_2 - \bm{C}}{\|\bm{A}_2 - \bm{C}\|},
\end{equation}
and the relative angular position of the candidate point is evaluated as
\begin{equation}
\theta
=
\arccos\!\left(\hat{\bm{u}}_1 \cdot \hat{\bm{u}}_2\right),\quad
\theta_c
=
\arccos\!\left(
\hat{\bm{u}}_1 \cdot
\frac{\bm{Q}_c - \bm{C}}{\|\bm{Q}_c - \bm{C}\|}
\right).
\end{equation}

If $0 \le \theta_c \le \theta$, the unconstrained solution $\bm{Q}_c$ lies within the arc span and is selected as the closest point. Otherwise, the closest endpoint of the arc is chosen. Accordingly, the closest point $\bm{Q}$ is given by
\begin{equation}
\bm{Q}
=
\begin{cases}
\bm{Q}_c, & 0 \le \theta_c \le \theta, \\[4pt]
\bm{A}_1, & \|\bm{P}_{obs} - \bm{A}_1\| \le \|\bm{P}_{obs} - \bm{A}_2\|, \\[4pt]
\bm{A}_2, & \text{otherwise (as $Q_2$ in Fig.~\ref{fig:collicheck})}.
\end{cases}
\end{equation}

Based on the resulting closest point, the signed safety distance is defined as
\begin{equation}
d_S
=
\|\bm{P}_{obs} - \bm{Q}\|
-
(\rho + r_{obs} + \delta),
\end{equation}
where $\rho$ denotes the cross-sectional radius of the continuum robot body, $r_{obs}$ is the obstacle radius, and $\delta > 0$ is a tunable safety buffer. Negative values of $d_S$ indicate a violation of the safety margin, corresponding to a collision state.

\begin{figure}[t]
    \centering    
    \includegraphics[width=0.8\linewidth]{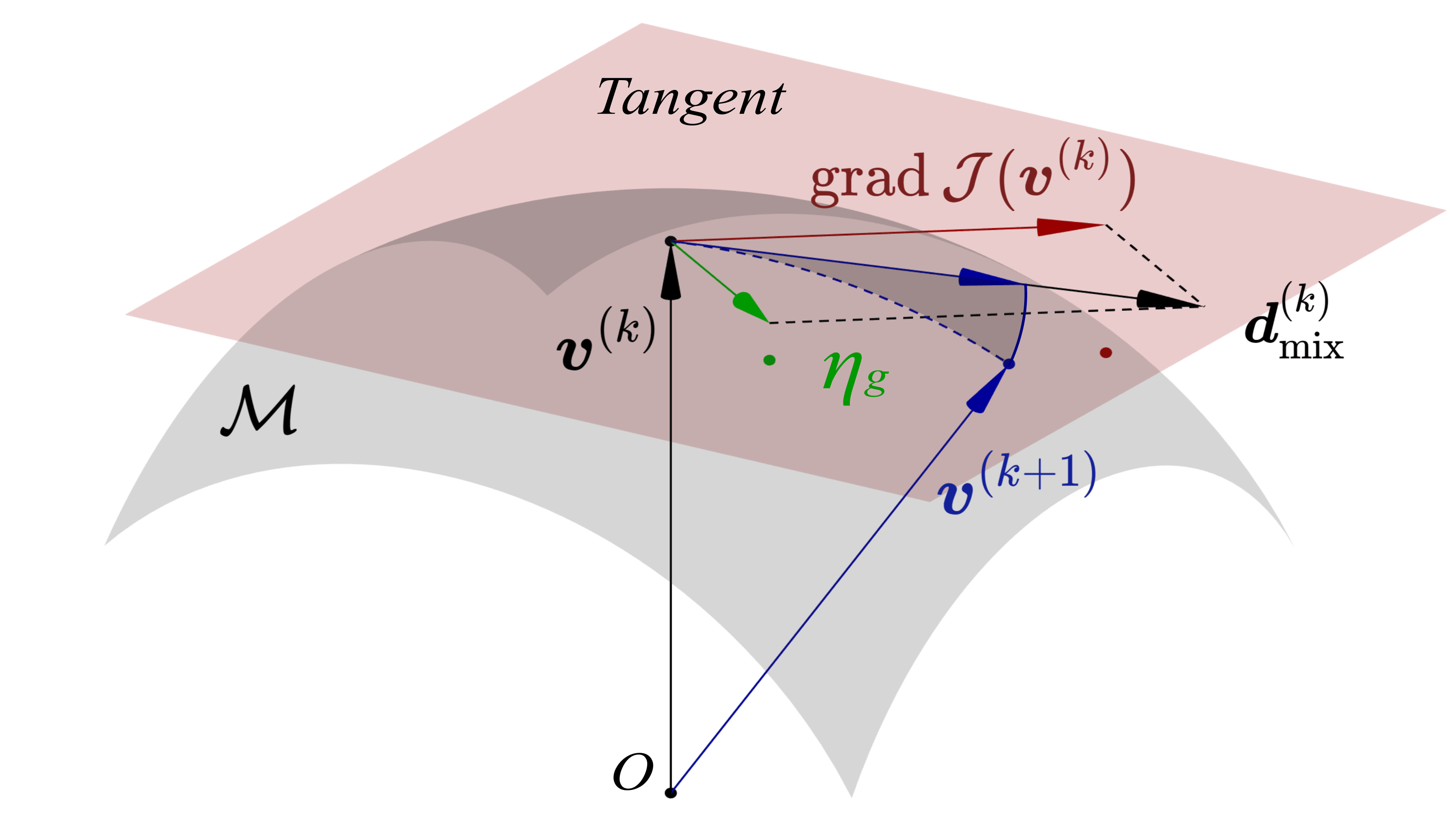}
    \caption{Illustration of the Riemannian optimization process with guided descent.}
    \label{fig:remanian}
\end{figure}

\subsection{Guided Directional Search}
\label{subsec:guided_search}

Efficient selection of search directions is critical for continuum robots operating in cluttered environments. Without explicit guidance, iterative search may perform redundant steps along directions that lead to collisions or oscillate near obstacles. Existing strategies often rely on heuristic biases or classical potential field methods, but these approaches either fail to ensure smooth, consistent guidance or do not fully exploit collision information.

To address these limitations, we propose a directional guidance mechanism that actively directs the search based on collision information. This mechanism generates a repulsive vector pointing away from nearby obstacles and projects it onto the current search direction to produce a smooth, feasible guidance vector. Unlike prior methods, our approach ensures consistent, collision-aware direction updates while accelerating convergence and mitigating oscillatory behavior.

Let the current search direction be denoted by \(\bm{t}_{\mathrm{c}} \in \mathbb{R}^3\), and the positions of obstacles be \(\bm{q}_{\mathrm{obs}}^{(i)}\), \(i=1,\dots,N\), we define a repulsive potential for each obstacle:
\begin{equation}
\mathcal{U}_{\mathrm{rep}}^{(i)}(\bm{t}_{\mathrm{c}}) =
\begin{cases}
\dfrac{1}{2} k_{\mathrm{rep}} \left( \dfrac{1}{\|\bm{t}_{\mathrm{c}} - \bm{q}_{\mathrm{obs}}^{(i)}\|} - \dfrac{1}{d_0} \right)^2, & \|\bm{t}_{\mathrm{c}} - \bm{q}_{\mathrm{obs}}^{(i)}\| < d_0 \\[1mm]
0, & \text{otherwise}
\end{cases}
\end{equation}
with the total potential
\begin{equation}
\mathcal{U}(\bm{t}_{\mathrm{c}}) = \sum_{i=1}^{N} \mathcal{U}_{\mathrm{rep}}^{(i)}(\bm{t}_{\mathrm{c}}) .
\end{equation}

The gradient of this potential provides the initial guidance vector:
\begin{equation}
\bm{F}_{\mathrm{g}} = - \nabla \mathcal{U}(\bm{t}_{\mathrm{c}}).
\end{equation}

To ensure smooth and feasible updates along the search, we project the guidance vector onto the subspace orthogonal to the current direction:
\begin{equation}
\bm{\eta}_{\mathrm{g}} = \frac{ \bm{P}_{\bm{t}_{\mathrm{c}}}^{\perp} \hat{\bm{f}}_{\mathrm{g}} }{ \left\| \bm{P}_{\bm{t}_{\mathrm{c}}}^{\perp} \hat{\bm{f}}_{\mathrm{g}} \right\| }, \quad
\bm{P}_{\bm{t}_{\mathrm{c}}}^{\perp} = \mathbb{I} - \bm{t}_{\mathrm{c}} \bm{t}_{\mathrm{c}}^\mathsf{T},
\label{eq:23}
\end{equation}
where \(\hat{\bm{f}}_{\mathrm{g}}\) is the normalized guidance vector. This projection guarantees that the corrected search direction is smooth, collision-aware, and aligned with the iterative search process.

By integrating collision information with directional correction, the proposed guided search efficiently biases the iterative exploration toward feasible directions, accelerates convergence, and produces smooth, collision-free trajectories, serving as a key enabler for the subsequent Riemannian manifold-based optimization.

\begin{algorithm}[t]
\caption{Guided Riemannian Optimization for Orientation Planning}
\label{alg:guided_rmo}
\begin{algorithmic}[1]
\Require Current orientation \(\bm{v}^{(0)}\), target \(\bm{p}_{\mathrm{target}}\), desired direction \(\bm{v}_{\mathrm{des}}\), step size \(\alpha\), guidance weight \(\lambda\), max iterations \(K_{\max}\)
\For{\(k = 0\) to \(K_{\max}-1\)}
    \State Compute forward kinematics \(\bm{p}_{\mathrm{ik}}(\bm{v}^{(k)})\)
    \State Evaluate Euclidean gradient \(\nabla \mathcal{J}_{\mathrm{total}}(\bm{v}^{(k)})\)
    \State Project gradient onto tangent space: \(\mathrm{grad}\,\mathcal{J}(\bm{v}^{(k)})\)
    \State Obtain guidance vector \(\bm{\eta}_{\mathrm{g}}^{(k)}\) from guided directional search
    \State Blend gradients: \(\bm{d}_{\mathrm{mix}}^{(k)} = (1-\lambda)(-\mathrm{grad}\,\mathcal{J}) + \lambda \bm{\eta}_{\mathrm{g}}^{(k)}\)
    \State Project mixed direction: \(\bm{d}^{(k)} = \mathbf{P}_{\bm{v}^{(k)}}^\perp \bm{d}_{\mathrm{mix}}^{(k)}\)
    \State Update orientation on manifold: \(\bm{v}^{(k+1)} = \exp_{\bm{v}^{(k)}}(\alpha \bm{d}^{(k)})\)
\EndFor
\State \Return Optimized orientation \(\bm{v}^{(K_{\max})}\)
\end{algorithmic}
\end{algorithm}

\subsection{Riemannian Optimization for Orientation Planning}
\label{subsec:riemannian_optimization}
Efficient orientation planning for continuum robotic manipulators requires maintaining both feasibility and smoothness while respecting physical constraints and collision avoidance. Conventional approaches often either neglect the geometric consistency of the manipulator or rely solely on gradient descent in Euclidean space, which can lead to infeasible or oscillatory updates when constrained to a unit-norm orientation vector. To address these limitations, we propose a guided Riemannian optimization framework that directly operates on the unit sphere manifold \(\mathbb{S}^2\), combining curvature-consistent inverse kinematics with a directional guidance mechanism. The guidance not only accelerates convergence but also prevents the optimizer from exploring infeasible orientations, thereby improving robustness and efficiency.

The proposed optimization framework achieves monotonic reduction in collision distance while maintaining computational efficiency throughout the iterative process. At iteration \(k\), the orientation of the end-effector is represented by a unit vector \(\bm{v}^{(k)} \in \mathbb{R}^3\), and the goal is to achieve alignment with a desired direction \(\bm{v}_{\mathrm{des}}\) while ensuring that the corresponding forward kinematics satisfy the target position \(\bm{p}_{\mathrm{target}}\). The total cost function \(\mathcal{J}(\bm{v}^{(k)})\) is defined as follows:

\begin{equation}
\begin{aligned}
\mathcal{J}(\bm{v}^{(k)}) &= w_{\mathrm{dir}} \underbrace{\left(1 - \langle \bm{v}^{(k)}, \bm{v}_{\mathrm{des}} \rangle \right)^2}_{\text{Directional alignment}} \\
&\quad + w_{\mathrm{pos}} \underbrace{\left\| \bm{p}_{\mathrm{ik}}(\bm{v}^{(k)}) - \bm{p}_{\mathrm{target}} \right\|^2}_{\text{Positional consistency}},
\end{aligned}
\end{equation}
where \(\langle \cdot, \cdot \rangle\) is the Euclidean inner product, \(w_{\mathrm{dir}}, w_{\mathrm{pos}} > 0\) are scalar weights reflecting the relative importance of directional versus positional objectives, and \(\bm{p}_{\mathrm{ik}}(\bm{v}^{(k)})\) is computed using the curvature-consistent IK solver described in Section~\ref{subsec:ik_solver_improved}. The directional term ensures that the orientation follows the desired pointing vector, while the positional term enforces geometric consistency of the manipulator's end-effector in Cartesian space.

For notational clarity, we introduce the auxiliary variables:
\begin{equation}
\left\{
\begin{aligned}
a^{(k)} &:= \langle \bm{v}^{(k)}, \bm{v}_{\mathrm{des}} \rangle, \\[2mm]
\Delta \bm{p}^{(k)} &:= \bm{p}_{\mathrm{ik}}(\bm{v}^{(k)}) - \bm{p}_{\mathrm{target}}, \\[1mm]
\mathbf{J}(\bm{v}^{(k)}) &:= \frac{\partial \bm{p}_{\mathrm{ik}}}{\partial \bm{v}} \Big|_{\bm{v} = \bm{v}^{(k)}},
\end{aligned}
\right.
\end{equation}
where \(\mathbf{J}(\bm{v}^{(k)})\) represents the Jacobian of the forward kinematics mapping from orientation space to Cartesian space. This Jacobian captures the sensitivity of the end-effector position to small changes in orientation, which is essential for accurate gradient-based updates.

The Euclidean gradient of the cost function is decomposed into two components corresponding to the directional and positional objectives:
\begin{equation}
\begin{aligned}
\nabla \mathcal{J}_{\mathrm{total}}(\bm{v}^{(k)}) &= \nabla \mathcal{J}_{\mathrm{dir}}(\bm{v}^{(k)}) + \nabla \mathcal{J}_{\mathrm{pos}}(\bm{v}^{(k)}), \\[1mm]
\nabla \mathcal{J}_{\mathrm{dir}}(\bm{v}^{(k)}) &= -2 w_{\mathrm{dir}} (1 - a^{(k)}) \bm{v}_{\mathrm{des}}, \\[1mm]
\nabla \mathcal{J}_{\mathrm{pos}}(\bm{v}^{(k)}) &= 2 w_{\mathrm{pos}}\, \mathbf{J}(\bm{v}^{(k)})^\top \Delta \bm{p}^{(k)}.
\end{aligned}
\end{equation}

Since \(\bm{v}^{(k)}\) must satisfy the unit-norm constraint, updates are constrained to the tangent space \(T_{\bm{v}^{(k)}} \mathbb{S}^2\) of the sphere at \(\bm{v}^{(k)}\). The projection operator is
\begin{equation}
\mathbf{P}_{\bm{v}^{(k)}}^\perp := \mathbf{I}_3 - \bm{v}^{(k)} \bm{v}^{(k)\top},
\end{equation}
and the Riemannian gradient is obtained as
\begin{equation}
\mathrm{grad}\, \mathcal{J}(\bm{v}^{(k)}) := \mathbf{P}_{\bm{v}^{(k)}}^\perp \nabla \mathcal{J}_{\mathrm{total}}(\bm{v}^{(k)}),
\end{equation}
which guarantees that the update direction remains tangent to the constraint surface, preserving feasibility at every iteration.

To enhance convergence and incorporate obstacle awareness, we introduce a guidance vector $\bm{\eta}_{\mathrm{g}}^{(k)}$ derived from the guided directional search in \eqref{eq:23}. This vector encodes information about safe and promising directions based on the manipulator’s current state and environment. The descent direction is then obtained by convexly blending the Riemannian gradient with the guidance:
\begin{equation}
\bm{d}_{\mathrm{mix}}^{(k)} = (1-\lambda)\,(-\mathrm{grad}\,\mathcal{J}(\bm{v}^{(k)})) + \lambda\, \bm{\eta}_{\mathrm{g}}^{(k)}, \quad \lambda \in [0,1],
\end{equation}
followed by projection back onto the tangent space:
\begin{equation}
\bm{d}^{(k)} := \mathbf{P}_{\bm{v}^{(k)}}^\perp \bm{d}_{\mathrm{mix}}^{(k)}.
\end{equation}

\begin{figure}[t]
    \centering    
    \includegraphics[width=0.85\linewidth]{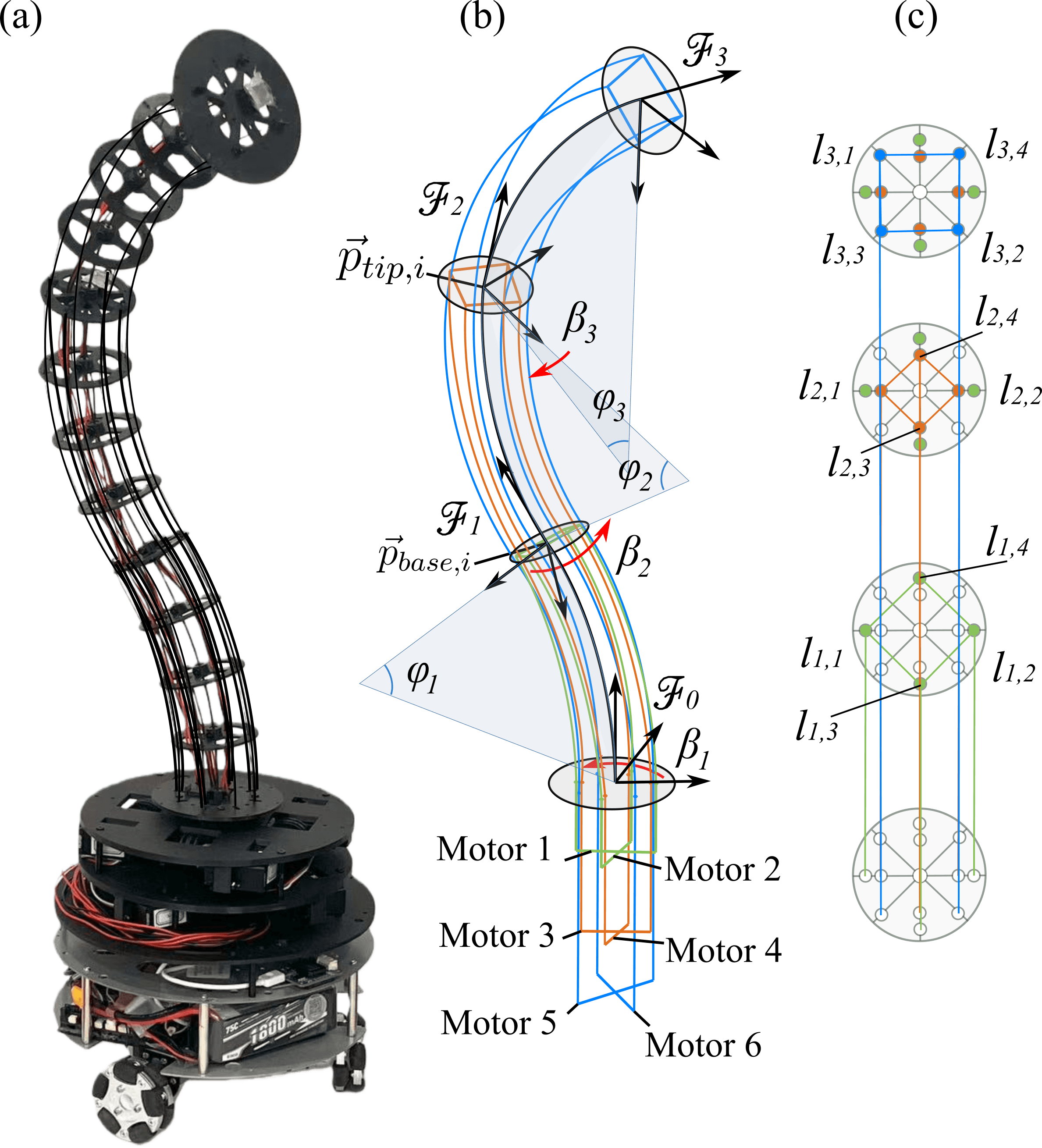}
    \caption{
    Structural overview and kinematic representation of the tendon-driven continuum robot.
    (a) Physical prototype of the mobile-base continuum robot.
    (b) Simplified tendon-driven backbone model showing segment-wise bending characterized by arc angle $\varphi_i$ and bending plane orientation $\beta_i$, with tendon routing and coordinate frames $\mathcal{F}_i$.
    (c) Tendon routing and structural layout for each segment, illustrating tendon paths through spacer disks and inter-segment connections.
    }
    \label{fig:config}
\end{figure}

Finally, the orientation is updated using the exponential map on \(\mathbb{S}^2\), which transports the tangent vector \(\bm{d}^{(k)}\) onto the manifold:
\begin{equation}
\bm{v}^{(k+1)} = \cos(\alpha \|\bm{d}^{(k)}\|)\, \bm{v}^{(k)} + \sin(\alpha \|\bm{d}^{(k)}\|)\, \frac{\bm{d}^{(k)}}{\|\bm{d}^{(k)}\|},
\end{equation}
where \(\alpha > 0\) is the step size. This update guarantees that \(\bm{v}^{(k+1)}\) remains a unit vector, maintaining both geometric and kinematic consistency.

Compared with traditional orientation optimization methods that may ignore curvature effects or collision guidance, our approach tightly couples curvature-consistent inverse kinematics with guided manifold updates. This integration improves convergence speed, ensures smooth orientation evolution across iterations, preserves physical plausibility of the manipulator shape, and prevents entrapment in infeasible orientations. The complete procedure is summarized in Algorithm~\ref{alg:guided_rmo}. The schematic overview in Fig.~\ref{fig:remanian} illustrates the overall guided Riemannian optimization process, highlighting the directional guidance, tangent-space projection, and manifold update steps.

\section{Geometry-Consistent FABRIK Solver with Curvature-Based Iteration}
\label{subsec:ik_solver_improved}
The classical FABRIK algorithm propagates constraints along the kinematic chain and enforces segment lengths without using Jacobians, but its assumption of straight rigid segments and focus on endpoint distances can lead to invalid shapes when applied to continuum robots \cite{aristidou2011fabrik}.

Some FABRIK variants for continuum robots approximate each continuum robot segment with two rigid links connected through a virtual joint \cite{Zhang2018, kolpashchikov2022}. Although this approximation enables the use of the FABRIK framework, it treats curvature as an implicit quantity derived from endpoint positions and ignores the intrinsic arc geometry of continuum segments. This may cause curvature discontinuities, shape distortion, and numerical instability.

To address these issues, we propose a new geometry-consistent FABRIK solver which preserves the backward–forward propagation structure of classical FABRIK but updates each segment in curvature space rather than in Euclidean joint space. Each intermediate configuration is thus guaranteed to correspond to a physically realizable circular-arc segment.

\begin{algorithm}[t]
\small
\caption{Geometry-Consistent FABRIK Solver with Curvature-Based Iteration for Continuum Robots}
\label{alg:fabrik_pcc_improved}
\begin{algorithmic}[1]
\Require 
Segment lengths $\{ l_i \}_{i=1}^n$, 
target position $\bm{p}_{\mathrm{target}}$, 
initial $\{ \kappa_i^{(0)}, \varphi_i^{(0)}, \beta_i^{(0)} \}_{i=1}^n$, 
convergence threshold $\tau$, max iterations $K_{\max}$
\Ensure 
Optimized parameters $\{ \kappa_i^*, \varphi_i^*, \beta_i^* \}$

\State Initialize $\bm{p}_{base,1} \gets \bm{0}$
\State Compute $\bm{p}_{tip,i}$ via forward mapping $\bm{F}(\mathbf{q}_i)$
\State $k \gets 0$
\While{ $\epsilon^{(k)} > \tau$ and $k < K_{\max}$ }
    \State \textbf{Backward phase:} set $\bm{p}_{tip,n} \gets \bm{p}_{\mathrm{target}}$
    \For{$i = n \text{ downto } 1$}
        \State $c_i \gets \|\bm{p}_{tip,i} - \bm{p}_{base,i}\|$
        \State Update $\kappa_i$ via Newton-Raphson on $\mathcal{R}_i(\kappa_i)$
        \State $\varphi_i \gets \kappa_i l_i + \Delta \varphi_i$
        \State $\bm{p}_{base,i} \gets \bm{p}_{tip,i} - \bm{F}(\mathbf{q}_i)$
    \EndFor
    \State \textbf{Forward phase:} $\bm{p}_{base,1} \gets \bm{0}$
    \For{$i = 1 \text{ to } n$}
        \State $\bm{p}_{tip,i} \gets \bm{p}_{base,i} + \bm{F}(\mathbf{q}_i)$
    \EndFor
    \State Update convergence metric $\epsilon^{(k)}$
    \State $k \gets k + 1$
\EndWhile
\State \Return $\{\kappa_i^*, \varphi_i^*, \beta_i^*\}$
\end{algorithmic}
\end{algorithm}

Each continuum segment is parameterized by
\begin{equation}
\mathbf{q}_i = [\kappa_i, \varphi_i, \beta_i]^T,
\end{equation}
where $\mathbf{q}_i$ defines the configuration of the $i$-th constant curvature segment. Specifically, $\kappa_i$ is the curvature magnitude, $\varphi_i$ is the total bending angle across the segment, and $\beta_i$ is the bending-plane direction, as shown in Fig.~\ref{fig:config}(b). The tip position of the $i$th segment is given by
\begin{equation}
\bm{p}_{tip,i} = \bm{p}_{base,i} + \frac{1}{\kappa_i}
\begin{bmatrix}
\sin \varphi_i \cos \beta_i \\
\sin \varphi_i \sin \beta_i \\
1 - \cos \varphi_i
\end{bmatrix}.
\end{equation}

\begin{figure*}[t]
    \centering   \includegraphics[width=0.8\linewidth]{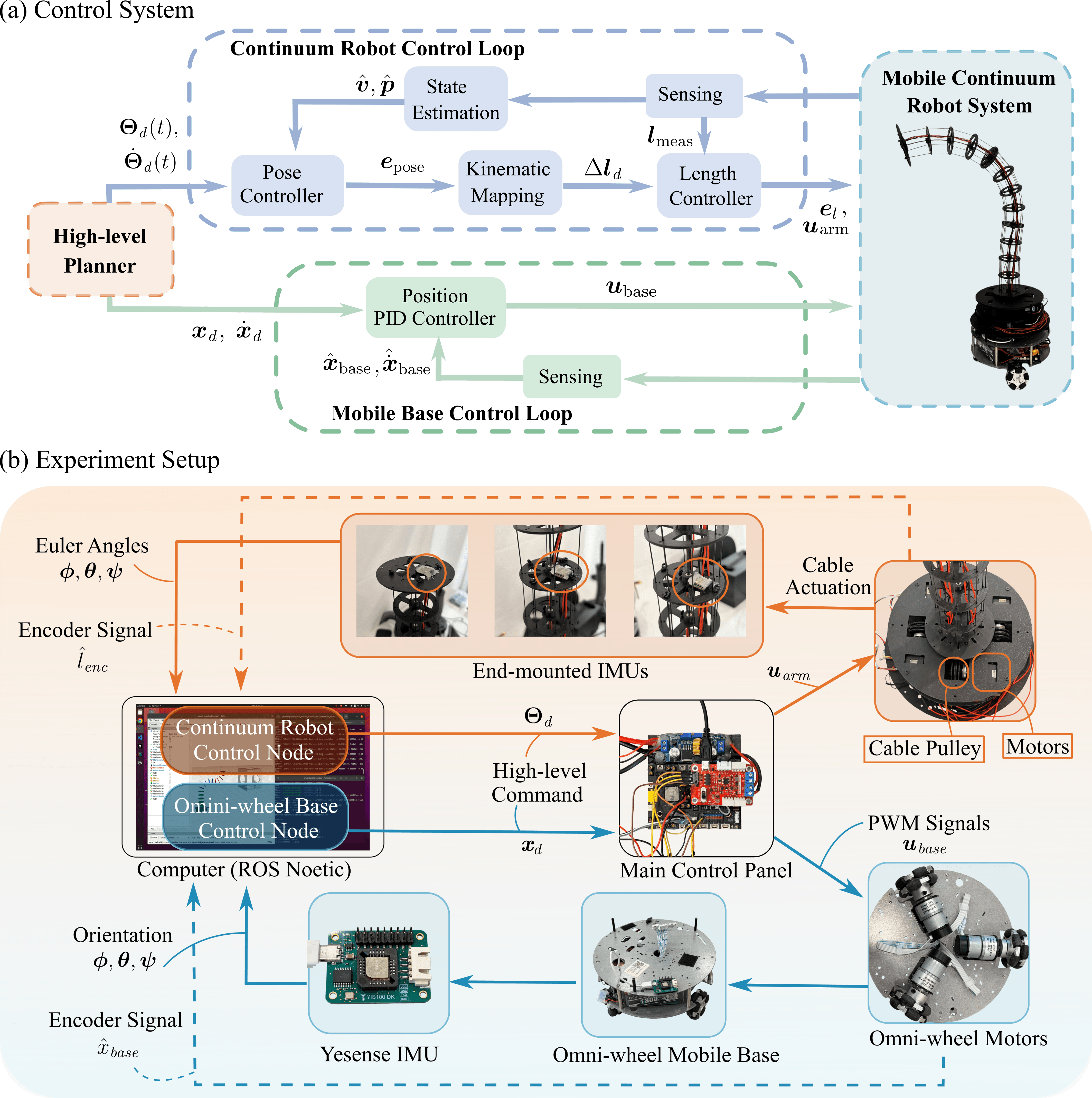}
    \caption{(a) Control architecture of the mobile continuum robot. The planner outputs manipulator and base trajectories, which are mapped to tendon length commands and base velocity references. Pose and cable-length errors drive the manipulator motors, while the position feedback regulates the mobile base. (b) Experimental Setup.}
    \label{fig:controller}
\end{figure*}

The arc angle is updated as
\begin{equation}
\varphi_i^{(k+1)} = \kappa_i^{(k)} l_i + \Delta \varphi_i^{(k)},
\end{equation}
where $\Delta \varphi_i^{(k)}$ adjusts the arc angle to maintain continuity during the propagation steps.

To ensure consistency between arc length and endpoint distance, curvature $\kappa_i$ is refined by solving the nonlinear geometric constraint
\begin{equation}
\mathcal{R}_i(\kappa_i) =
2\frac{1}{\kappa_i}
\arcsin\!\left(\frac{c_i\kappa_i}{2}\right) - l_i = 0,
\end{equation}
where $c_i = \|\bm{p}_{tip,i} - \bm{p}_{base,i}\|$ is the chord length of the segment. The curvature $\kappa_i$ is updated using the Newton–Raphson method:

\begin{align}
\kappa_i^{(k+1)} &= \kappa_i^{(k)} -
\frac{\mathcal{R}_i(\kappa_i^{(k)})}{
\partial \mathcal{R}_i / \partial \kappa_i}, \\
\frac{\partial \mathcal{R}_i}{\partial \kappa_i} =& 
- \frac{2}{\kappa_i^2}
\arcsin\!\left(\frac{c_i\kappa_i}{2}\right)
+ \frac{c_i/\kappa_i}{
\sqrt{1-(c_i\kappa_i/2)^2}}.
\end{align}

This curvature-based update avoids the numerical difficulty associated with radius-based formulations in near-straight configurations.

The forward–backward propagation steps follow the FABRIK procedure but operate on circular-arc segments:
\begin{align}
\bm{p}_{base,i}^{(k+1)} &= \bm{p}_{tip,i}^{(k)} - \bm{F}(\mathbf{q}_i^{(k)}), \\
\bm{p}_{tip,i}^{(k+1)} &= \bm{p}_{base,i}^{(k+1)} + \bm{F}(\mathbf{q}_i^{(k+1)}),
\end{align}
where $\bm{F}(\mathbf{q}_i)$ denotes the forward kinematics of a constant-curvature arc.

Convergence is evaluated using a combined position–curvature metric:
\begin{equation}
\epsilon^{(k)} =
\max_i \left(
\|\bm{p}_{tip,i}^{(k)} - \bm{p}_{tip,i}^{(k-1)}\|,
\,
|\kappa_i^{(k)} - \kappa_i^{(k-1)}|
\right).
\end{equation}

The overall algorithm is summarized in Algorithm~\ref{alg:fabrik_pcc_improved}.

\section{Low-level Motion Controller}
\label{sec:control}

A dual-loop sensor-driven controller tracks the planned continuum configuration trajectory, while a PID controller regulates the mobile base using the desired configuration, cable lengths, and base commands derived from the planner, as shown in Fig.~\ref{fig:config}.

The manipulator control is organized as an outer-loop pose controller and an inner-loop cable-length controller. The outer-loop computes the orientation and position error:
\begin{equation}
\boldsymbol{e}_\text{pose} =
\begin{bmatrix}
\boldsymbol{v}_\mathit{d} - \hat{\boldsymbol{v}} \\
\boldsymbol{p}_\mathit{d} - \hat{\boldsymbol{p}}
\end{bmatrix},
\end{equation}
and generates the cable-length reference increment \(\Delta \boldsymbol{l}_\mathit{d}\), which is combined with nominal lengths \(\boldsymbol{l}_\text{nominal}\) to form the inner-loop reference
\begin{equation}
\boldsymbol{l}_\text{ref} = \boldsymbol{l}_\text{nominal} + \Delta \boldsymbol{l}_\mathit{d}.
\end{equation}
The inner-loop cable-length controller computes the actuator commands based on the measured cable lengths \(\boldsymbol{l}_\text{measured}\):
\begin{equation}
\begin{aligned}
\bm{e}_l &= \bm{l}_\mathit{ref} - \bm{l}_\mathit{measured}, \\[4pt]
\bm{u}_\mathit{arm} &= \mathbf{K}_p \bm{e}_l + \mathbf{K}_d \dot{\bm{e}}_l
+ \mathbf{K}_\mathit{ff} \dot{\bm{\Theta}}_\mathit{d}
+ \mathbf{K}_\mathit{pose} \bm{e}_\mathit{pose}.
\end{aligned}
\end{equation}

State estimation combines encoder and IMU measurements \(\boldsymbol{l}_\text{measured}, \boldsymbol{v}_\text{IMU}\) to provide the current end-effector position \(\hat{\boldsymbol{p}}\), orientation \(\hat{\boldsymbol{v}}\), and configuration \(\hat{\boldsymbol{\Theta}}\), which are used by the outer-loop controller.

The mobile base is controlled via a standard PID loop using odometry and wheel encoder feedback \(\hat{\boldsymbol{x}}_\text{base}, \hat{\dot{\boldsymbol{x}}}_\text{base}\), generating base motor commands \(\boldsymbol{u}_\text{base}\) to track \(\boldsymbol{x}_\mathit{d}, \dot{\boldsymbol{x}}_\mathit{d}\).
This multi-loop, sensor-driven design, illustrated in Fig.~\ref{fig:controller}(a), enables the mobile continuum robot to robustly and precisely track both manipulator and base trajectories in constrained environments.

\section{Simulation and Experimental Results}
\label{sec:results}

This section validates the proposed motion planning and control framework. After describing the experimental setup, we present comparative simulations in a pipeline environment against several optimization-based planners, focusing on computational efficiency and planning quality. Real-world experiments of increasing difficulty are then conducted, progressing from basic obstacle avoidance and shape adaptation on a fixed-base platform to coordinated base–arm motion in dynamic, cluttered environments.

\subsection{Experimental Setup}

The platform comprises a cable-driven continuum manipulator and a mobile base, as shown in Fig.~\ref{fig:controller}(b). An industrial PC with an Intel Core i7 processor (3.0 GHz) and 16 GB RAM runs Ubuntu 20.04 and ROS Noetic for high-level planning, communicating with a DJI RoboMaster A control board via a serial interface. Feetech STS3046 servos actuate the manipulator, while MG513 motors drive the mobile base. Manipulator feedback is provided by segment-mounted IMUs and motor encoders, and base motion is regulated by PID control using IMU and wheel-odometry feedback.

\begin{figure}[t]
    \centering
    \includegraphics[width=0.85\linewidth]{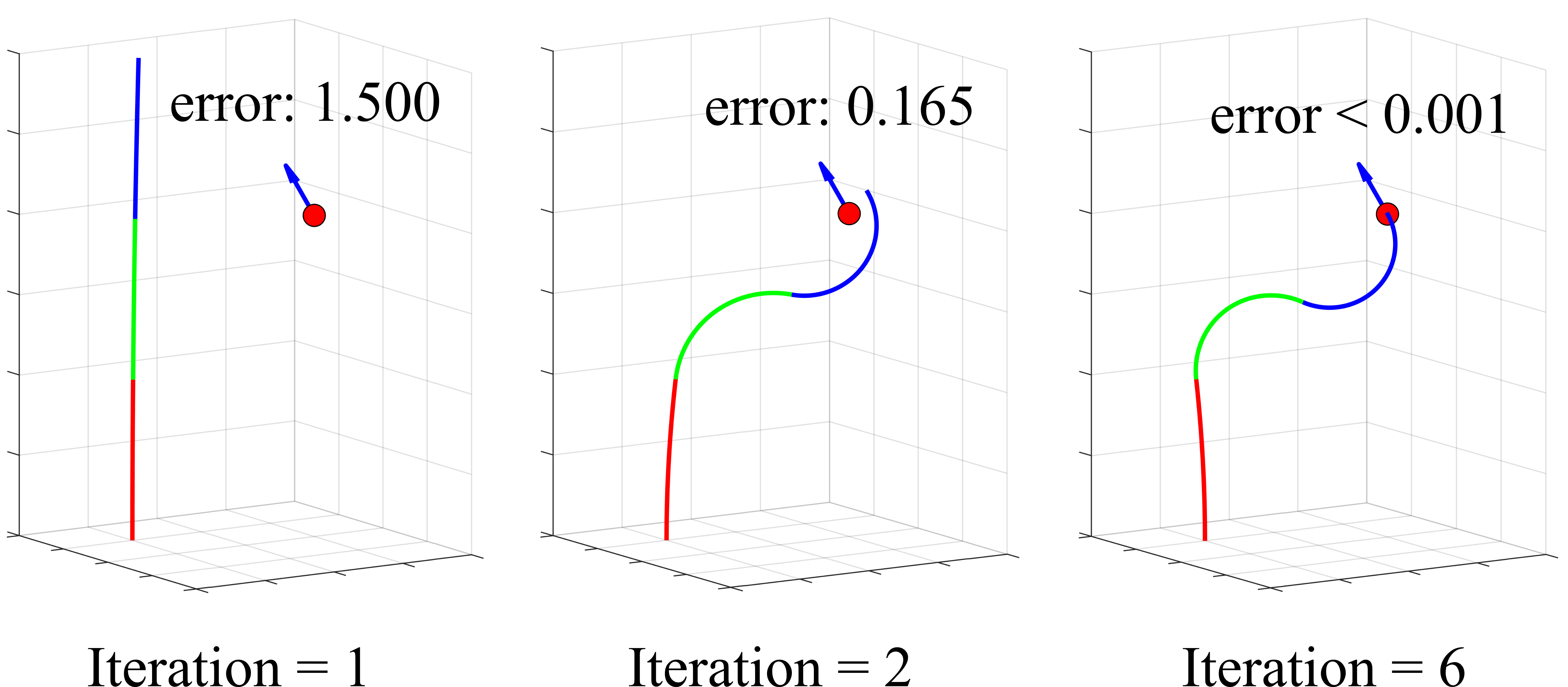}
    \caption{Iterative inverse kinematics process illustrating how the end-effector converges to the target. The initial upright pose (iteration = 1) starts with a large error. After the first iteration (iteration = 2), orientation is aligned while position error remains. Subsequent iterations refine the solution until convergence at iteration~6 with position error below 0.001 (unit link length).}

    \label{fig:ik_iter}
\end{figure}

\begin{figure}[t]
    \centering
    \includegraphics[width=0.8\linewidth]{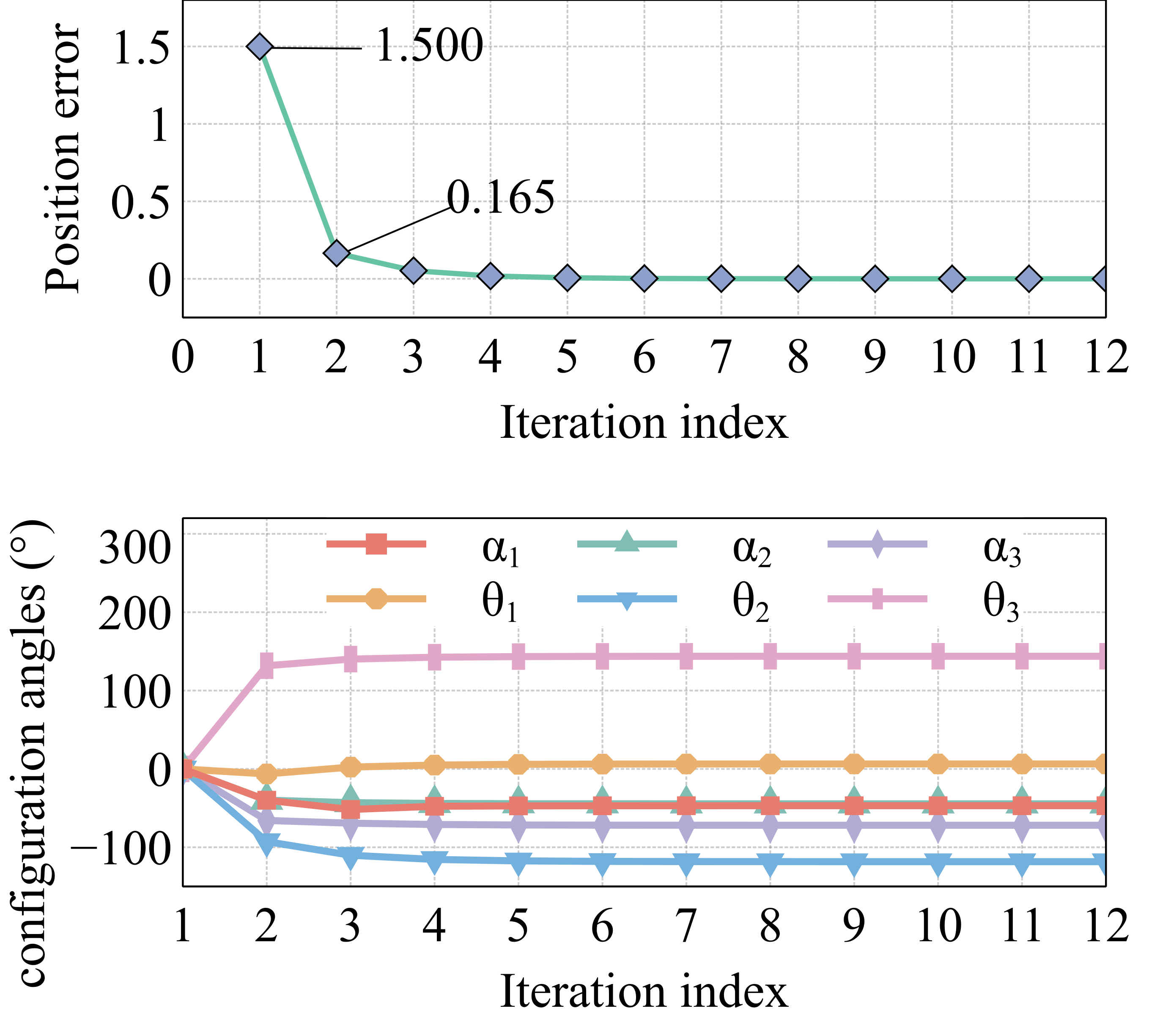}
    \caption{Convergence of the inverse kinematics solution. Top: position error (unit link length) versus iteration, showing a rapid drop from 1.5 to 0.165 in the first iteration and convergence below 0.001 thereafter. Bottom: evolution of configuration angles, indicating that a near-optimal solution is already reached at iteration~2.}

    \label{fig:ik_convergence}
\end{figure}

\begin{figure*}[t]
    \centering   \includegraphics[width=0.9\linewidth]{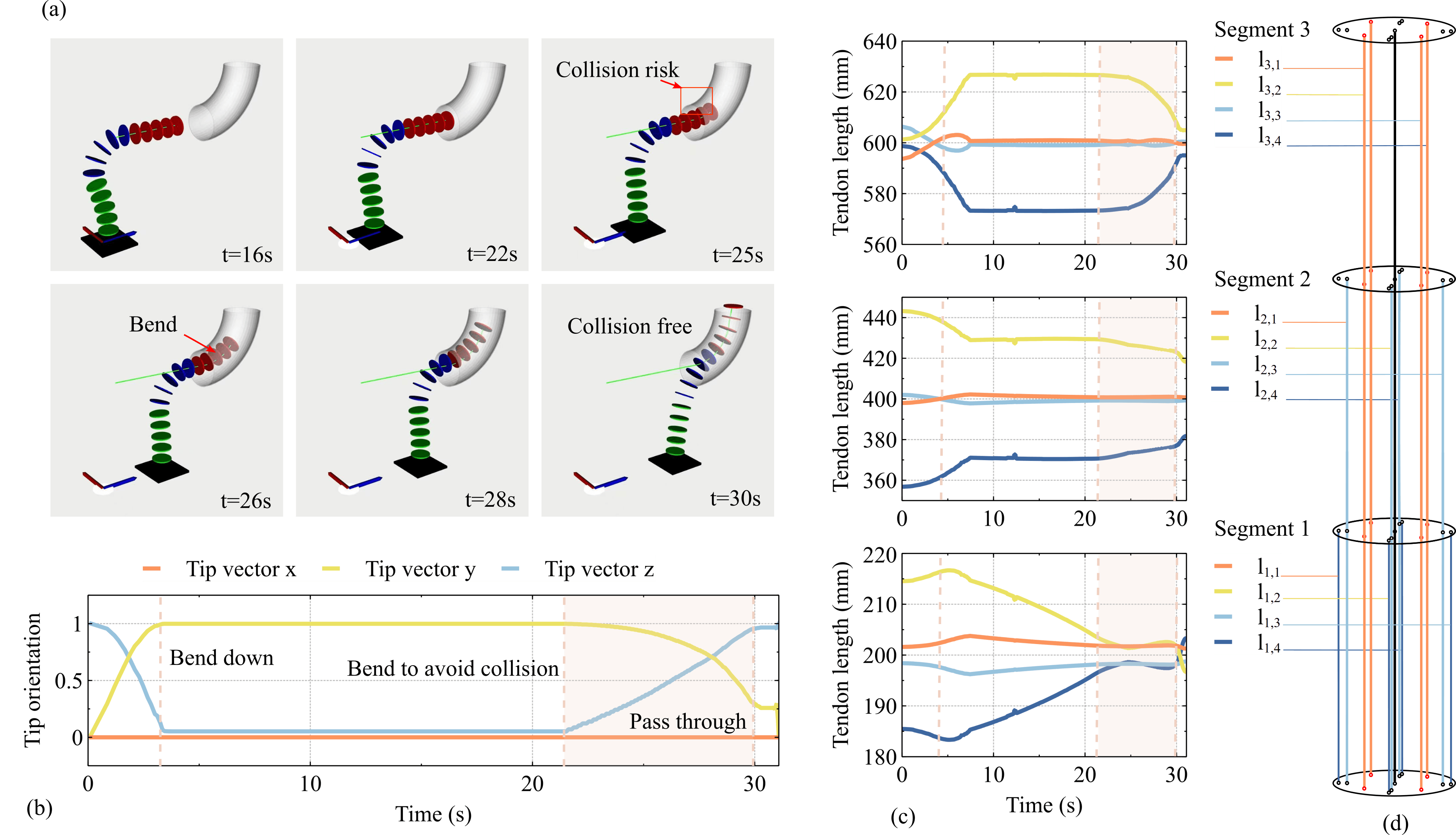}
    \caption{Pipe traversal simulation of the mobile continuum robot: (a) Snapshots of the robot navigating through the pipe, (b) Evolution of the end-effector orientation vector over time, (c) Tendon length variations during motion, and (d) Schematic of the three-section continuum robot's tendon configuration.}
    \label{fig:simulation1}
\end{figure*}

\begin{table}[t]
\centering
\footnotesize
\caption{Performance Comparison of Inverse Kinematics Solvers}
\begin{tabular}{ m{1cm}  m{1cm}  m{1cm}  m{1cm}  m{1cm}  m{1cm} }
\toprule
Method & Avg Time (ms) & Max Time (ms) & Avg Iter & Max Iter & Success Rate (\%)\\
\midrule 
\textbf{Ours} & \textbf{2.20} & \textbf{21.84} & \textbf{4.16} & \textbf{8} & \textbf{100} \\
\midrule
FABRIKc \cite{Zhang2018} & 22.2 & 82.6 &44.55 &112 & -- \\
\midrule
FABRIKx \cite{kolpashchikov2022} & 3.3 & -- & 17 & -- & 93.8\\
\midrule
Jacobian \cite{Zhang2018} &192.6 & 217.7& 1604.8 &1722 & --  \\
\bottomrule
\end{tabular}
\label{tab:ik_solver_comparison}
\end{table}

\subsection{Performance Evaluation of the Geometry-Consistent FABRIK Solver}

 In this subsection, we evaluate the proposed geometry-consistent FABRIK solver through numerical experiments conducted on a three-segment continuum robot with unit link lengths. As shown in Fig.~\ref{fig:ik_iter}, the end-effector converges to the target within six iterations starting from an upright configuration. Specifically, as shown in Fig.~\ref{fig:ik_iter} and the upper plot of Fig.~\ref{fig:ik_convergence}, after the first iteration (iteration~2), the position error drops sharply from 1.5 to 0.165. The final position error is below $10^{-3}$, and the configuration parameters exhibit fast stabilization, indicating rapid and stable convergence behavior.

A quantitative comparison with existing methods is reported in Table~\ref{tab:ik_solver_comparison} based on a 100-point trajectory tracking task. The proposed method achieves an average computation time of 2.20~ms per IK query, which is significantly faster than FABRIKc and the Jacobian-based solver \cite{Zhang2018}. Meanwhile, it requires only 4.16 iterations on average, compared to tens or even thousands for the other methods \cite{Zhang2018,kolpashchikov2022}. These results confirm that directly optimizing in curvature space leads to superior computational efficiency and robustness.

\begin{figure}
    \centering   \includegraphics[width=0.8\linewidth]{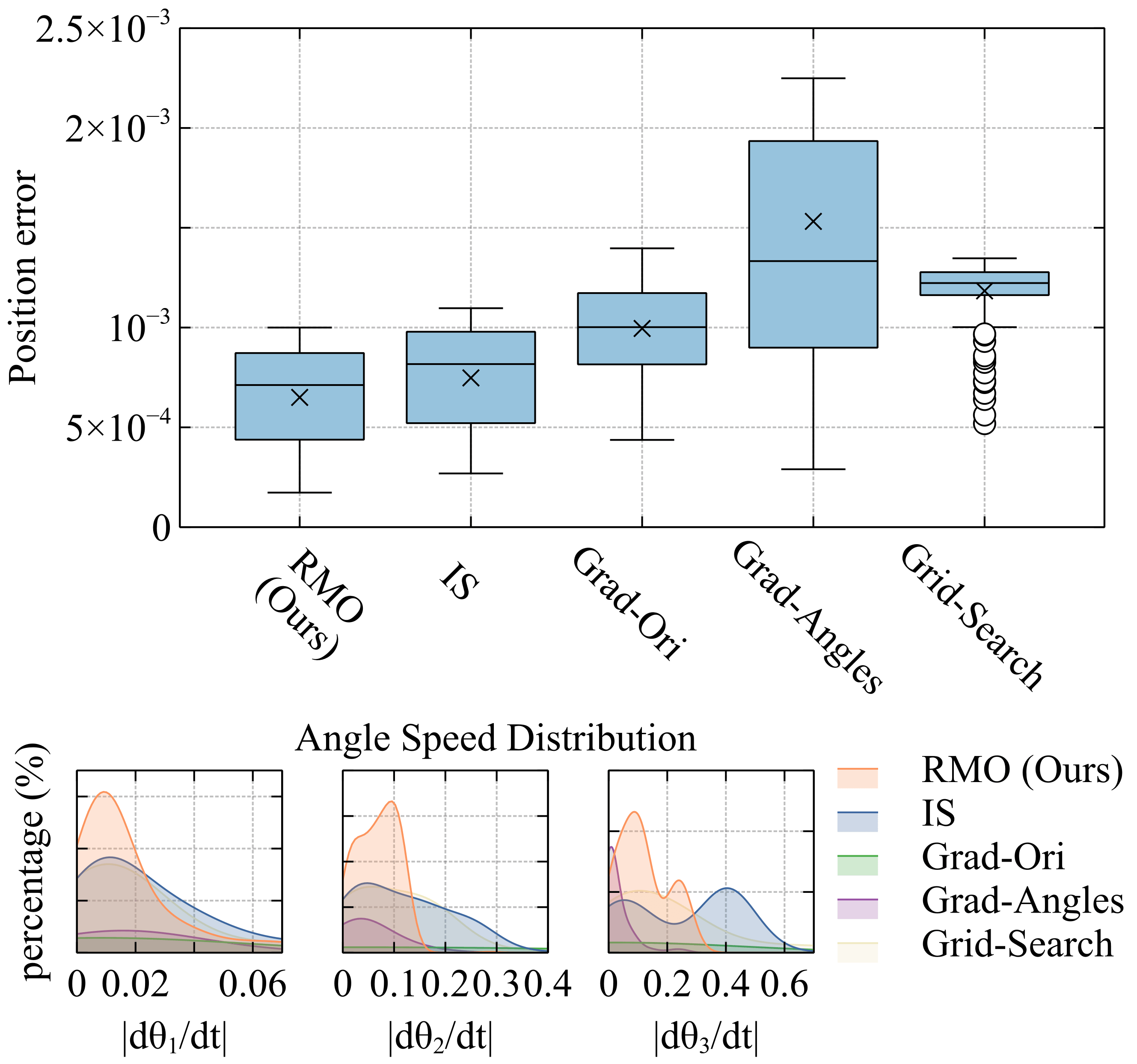}
    \caption{Quantitative comparison of our method and baseline planning algorithms. The upper plot presents the position error comparison (normalized by the continuum link length of 1.5 units). The lower plot presents the distribution of the absolute values of angular velocity changes, which serves as the primary metric for trajectory smoothness.}
    \label{fig:simulation2}
\end{figure}

\begin{table}
\centering
\footnotesize
\caption{Quantitative Comparison of Planning Methods} % 标题更具学术性
\begin{tabular}{ m{0.8cm}  m{2.4cm}  m{1.0cm}  m{0.6cm}  m{0.6cm}  m{0.6cm} } % 调整了列宽，以适应更长的描述
\toprule
Method & Optimization Strategy & Search Space & Avg Time (ms) & Avg Iterations & Success Rate (\%) \\
\midrule
\textbf{RMO (Ours)} & \textbf{Geometric Heuristic (with Riemannian Manifold Opt.)} & \textbf{Ori} & \textbf{6} & \textbf{15} & \textbf{100} \\
\midrule
IS (without RMO) & Geometric Heuristic (without Manifold Geometry) & Ori & 14 & 32 & 98 \\
\midrule
Grad-Ori & Euclidean gradient optimization & Ori & 43 & 64 & 80 \\
\midrule
Grad-Angles & Euclidean gradient optimization & Config & 253 & 79 & 62 \\
\midrule
Grid-Search & Exhaustive discrete search & Ori & 132 & 75 & 55 \\
\bottomrule
\end{tabular}
\label{tab:baseline_comparison}
\end{table}

\begin{figure*}[t]
    \centering
    \includegraphics[width=0.85\linewidth]{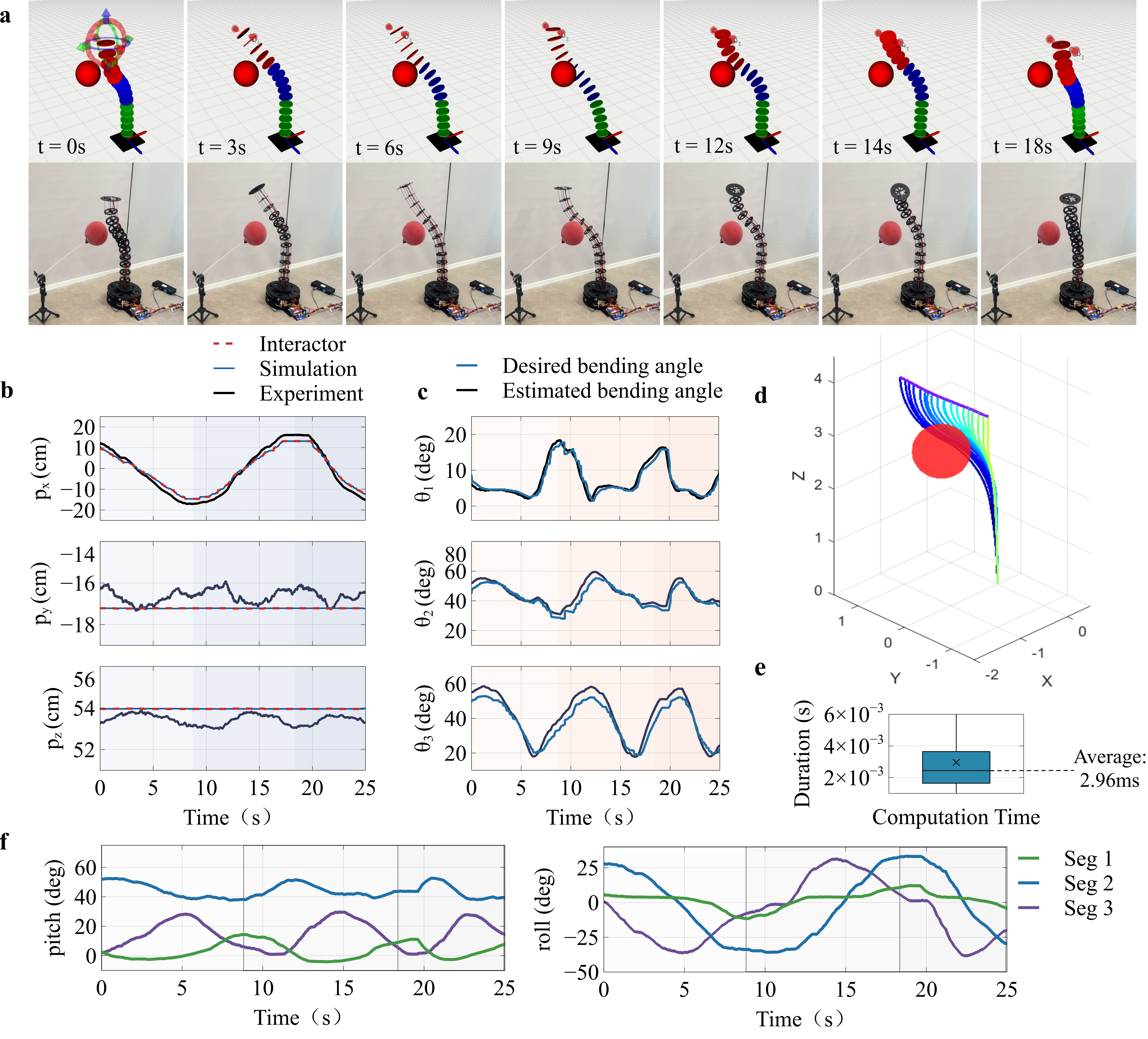}
    \caption{End-Effector Control with Obstacle Avoidance. (a) Motion snapshots: simulation (top), physical (bottom); (b) End-effector trajectories (reference via RViz marker, simulated, experimental); (c) Bending angle tracking; (d) Arm configuration evolution; (e) Computation time box-plot; (f) End-effector pitch/roll angles measurement.}
    \label{fig:experiment1}
\end{figure*}

\begin{figure}
    \centering
    \includegraphics[width=0.95\linewidth]{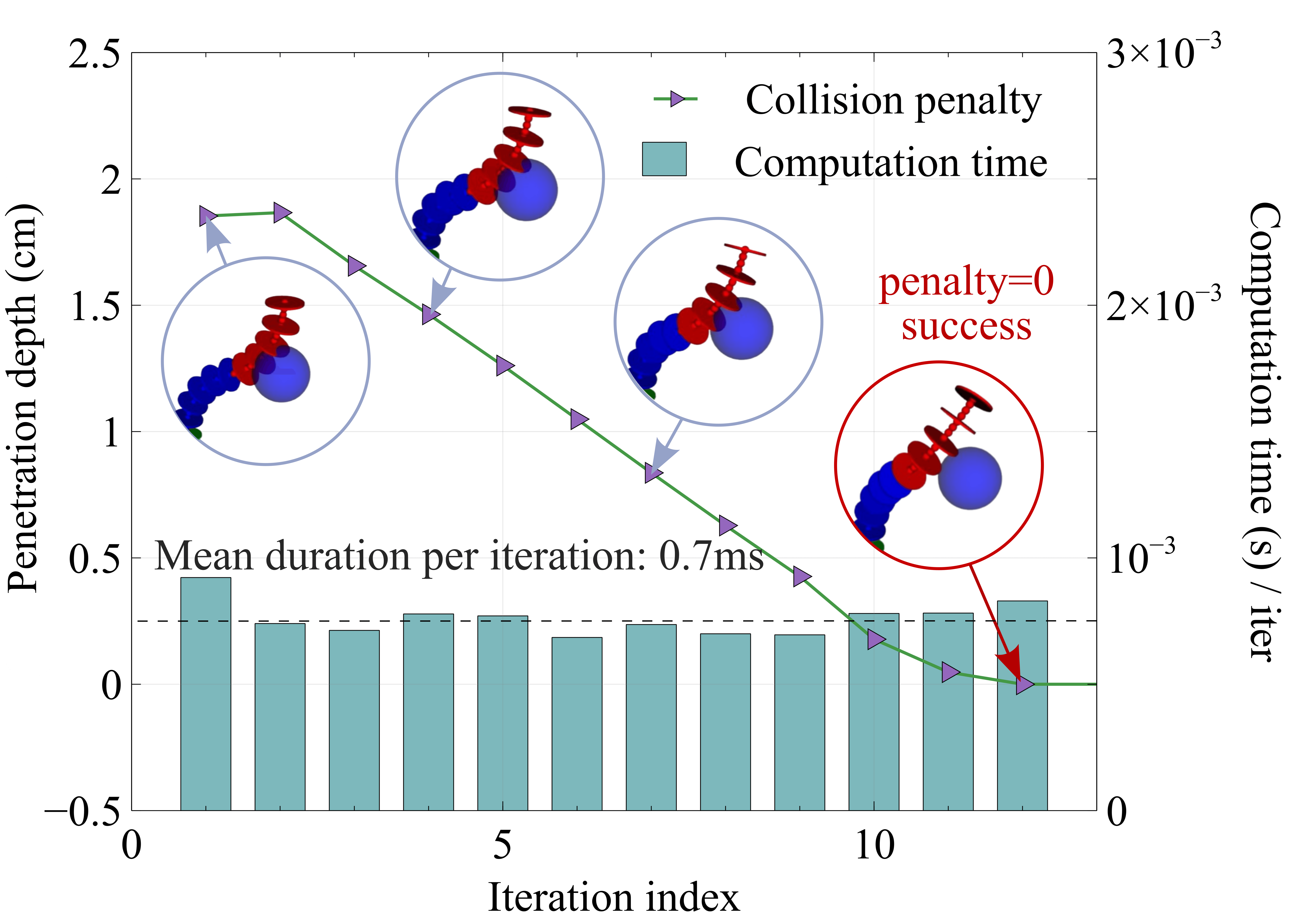}
    \caption{Evolution of penetration depth and computation time over iterations. The curve (left y-axis) shows the penetration depth decreasing until a collision-free state is achieved. The bar graph (right y-axis) represents the computation time per iteration.}
    \label{fig:avoid_process}
\end{figure}

\subsection{Simulation-based Evaluation and Baseline Comparison}
\label{subsec:simulation}

We evaluate the proposed framework in a simulated pipe-traversal task and compare its planning performance against four baseline planners. As shown in Fig.~\ref{fig:simulation1}(a), the omnidirectional base and continuum manipulator coordinate their motion to traverse the pipeline without collisions. Fig.~\ref{fig:simulation1}(b) and (c) show the end-effector orientation and tendon length profiles, respectively, while Fig.~\ref{fig:simulation1}(d) illustrates the tendon routing structure.

Table~\ref{tab:baseline_comparison} compares RMO with IS and three benchmark planners. IS serves as an ablation baseline using step-based search instead of Riemannian optimization. Grad-Ori applies gradient-based optimization directly in orientation space \cite{absil2008optimization}, whereas Grad-Angles optimizes in configuration angle space \cite{ratliff2009chomp}. Grid-Search exhaustively evaluates a discretized orientation space.

These comparisons assess computational efficiency and trajectory quality, examining the effects of the optimization space and the Riemannian formulation.

\begin{figure*}[t]
    \centering    \includegraphics[width=0.9\linewidth]{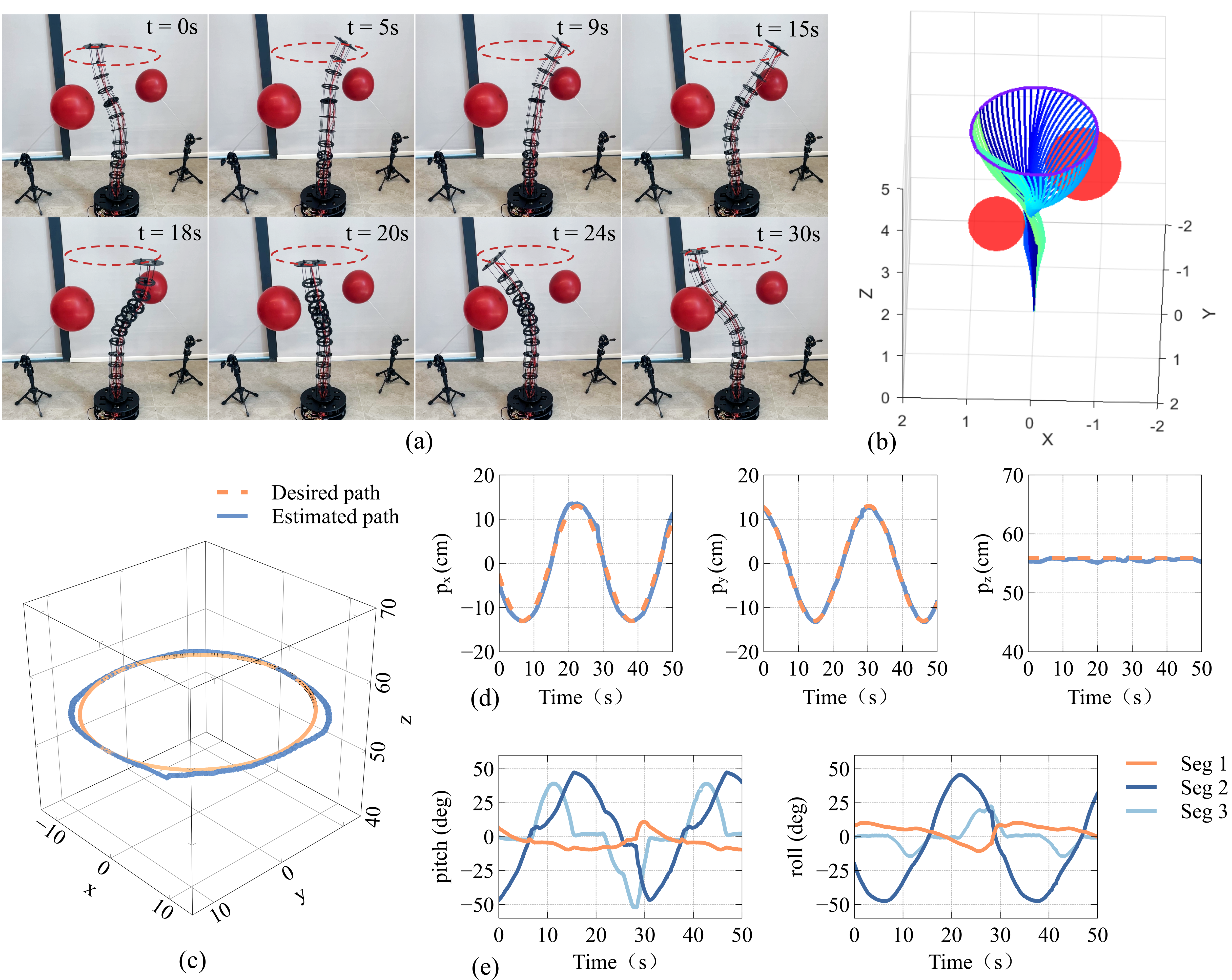}
    \caption{Experiment 2: Circular trajectory tracking experiment with two spherical obstacles. (a) Snapshots of the fixed-base continuum robot tracking a circular path while avoiding obstacles, (b) Arm configuration evolution during motion, (c) 3D end-effector trajectory (desired vs. estimated), (d) Position coordinates $(P_x, P_y, P_z)$ over time, (e) Euler angle variations.}
    \label{fig:experiment2}
\end{figure*}

\begin{figure*}[t]
    \centering    \includegraphics[width=0.9\linewidth]{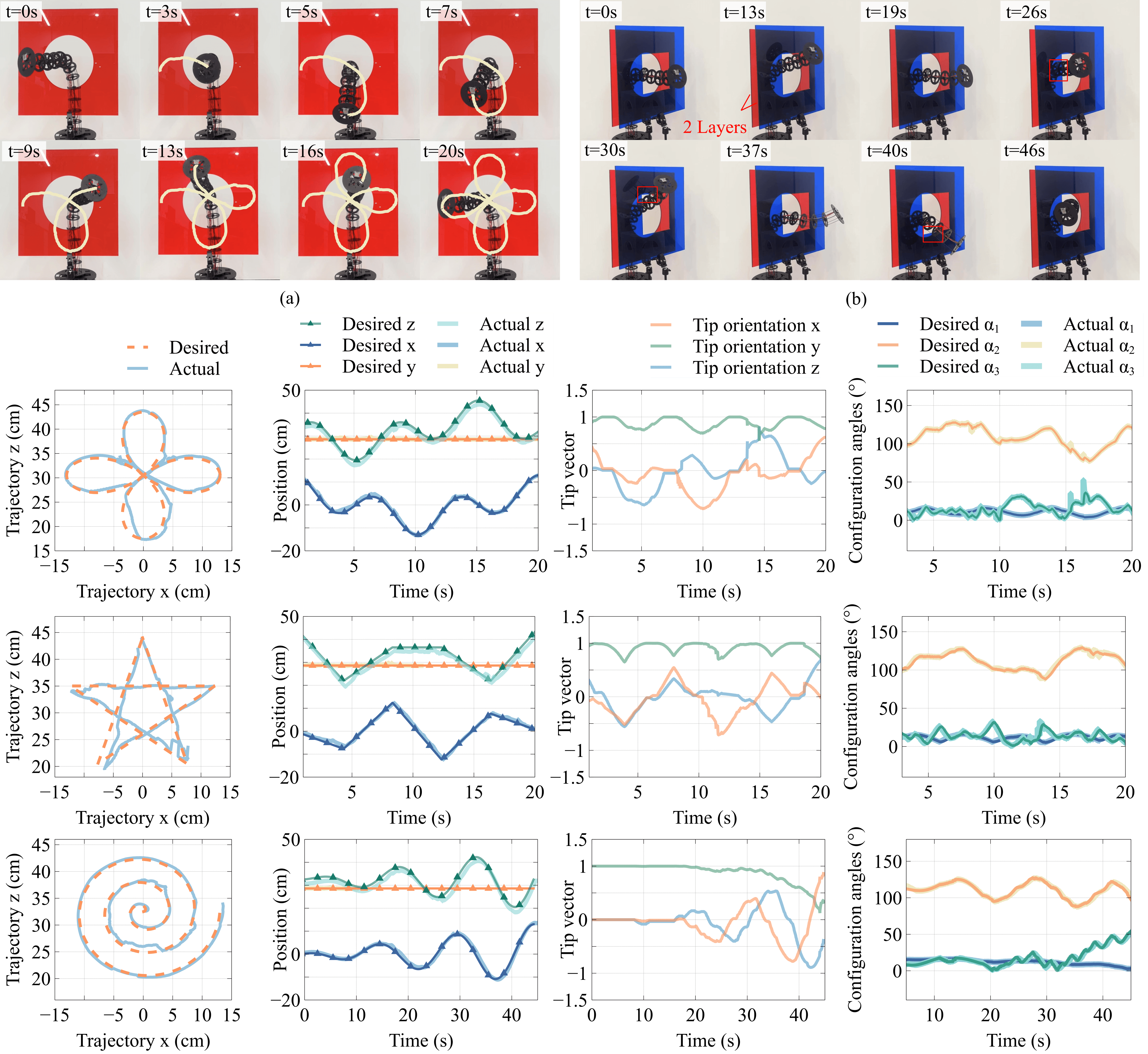}
    \caption{Experiment 3: Trajectory tracking through barrier plates. Top: Snapshots of (a) four-leaf clover, (b) pentagram, and (c) helical trajectories. Bottom: $3 \times 4$ subplots showing quantitative performance; first two rows correspond to single barrier plate trajectories, third row corresponds to the helical trajectory with dual-layer barrier plates.}
    \label{fig:experiment3}
\end{figure*}

Quantitative comparisons are presented in Fig.~\ref{fig:simulation2} and Table~\ref{tab:baseline_comparison}, confirming the superior performance of our RMO method across all key metrics. The upper plot in Fig.~\ref{fig:simulation2}, illustrating position errors (normalized by the continuum link length of 1.5 units), shows that RMO consistently achieves the lowest error metrics, reflecting superior spatial accuracy and overall robustness. Furthermore, Table~\ref{tab:baseline_comparison} confirms that RMO maintains the highest obstacle avoidance success rate under the 100-iteration limit while demonstrating competitive planning speed against the fastest local optimizers.

The lower plot in Fig.~\ref{fig:simulation2} presents angular velocity changes, serving as the primary metric for trajectory smoothness. The distribution of RMO is highly concentrated around zero, indicating minimal movement corrections and resulting in superior smoothness. In contrast, comparison planners show dispersed distributions, with substantial presence at higher absolute velocity values. The IS method, for instance, exhibits large, coarse velocity variations due to its discrete steps. The gradient-based methods (Grad-Ori and Grad-Angles) similarly show occasional abrupt spikes, confirming their susceptibility to local minima.

\subsection{Obstacle Avoidance and Shape Adaptation Experiments with Fixed Base}

\label{subsec:fixed_base}

Building upon the simulation results that validate the algorithmic core of our framework, we now focus on its real-world performance. The following physical experiments are designed to evaluate the complete system under increasingly complex conditions.

The first group of experiments verifies the fundamental capability of the proposed framework using a fixed-base continuum robot. Both simulations and physical experiments are conducted to assess real-time collision avoidance and shape adaptation performance.

\subsubsection{End-Effector Control Experiment with Obstacle Avoidance}

As illustrated in Fig.~\ref{fig:experiment1}(a), a comparative sequence of snapshots from both simulation and physical experiments captures the motion of the manipulator from $t = 0$ to $18$ \si{\second}. The end-effector motion was commanded manually via the interactive marker in RViz, allowing direct 6-DOF control of the tip pose. At the initial state ($t = 0$ \si{\second}), the manipulator maintains an upward end-effector orientation. By $t = 3$ \si{\second}, a potential collision risk emerges between the second and third segments and the obstacle (red sphere), prompting reactive bending away from the obstacle for safe passage. During $t = 6$--$9$ \si{\second}, as the commanded end-effector motion directs it away from the obstacle, the collision risk diminishes and the end-effector orientation gradually recovers upward. Throughout the return phase ($t = 9$--$18$ \si{\second}), the manipulator consistently adapts its shape near the obstacle, successfully maintaining both safety and tracking accuracy.

Fig.~\ref{fig:experiment1}(b) illustrates the end-effector position trajectories ($p_x$, $p_y$, $p_z$) versus time, comparing the desired (user-commanded via RViz interactive marker), simulated, and experimental paths. The simulated trajectory exhibits close alignment with the reference commands, while the physical robot accurately tracks the reference during reciprocating motion along the x-axis ($-20$ to $20$ \si{\centi\meter}), with unidirectional traversal requiring approximately $8$ \si{\second}.

The corresponding configuration-space behavior appears in Fig.~\ref{fig:experiment1}(c), showing the temporal evolution of bending angles $\theta_1$, $\theta_2$, $\theta_3$. Both desired angles (computed by the obstacle avoidance module based on the commanded end-effector pose) and estimated physical angles demonstrate consistent deformation control throughout operation.

Fig.~\ref{fig:experiment1}(d) visually emphasizes the manipulator's shape adaptation during obstacle avoidance, highlighting its dynamic reconfiguration capability in constrained workspaces. Computational efficiency of the obstacle avoidance algorithm is quantified in Fig.~\ref{fig:experiment1}(e) via box-plot analysis. With a mean computation time of \SI{2.96} {\milli\second}, the approach satisfies real-time control requirements. The computational efficiency of this optimization process is further reflected in the computation time profile shown in Fig.~\ref{fig:avoid_process}, where the per-iteration cost remains consistently low throughout the iterations, while the penetration depth (left y-axis) decreases monotonically until a collision-free configuration is achieved. Finally, Fig.~\ref{fig:experiment1}(f) presents pitch and roll angles measured by IMU sensors at the physical manipulator's end-effector, confirming smooth motion execution without abrupt orientation changes.

\begin{figure*}[t]
    \centering   \includegraphics[width=0.9\linewidth]{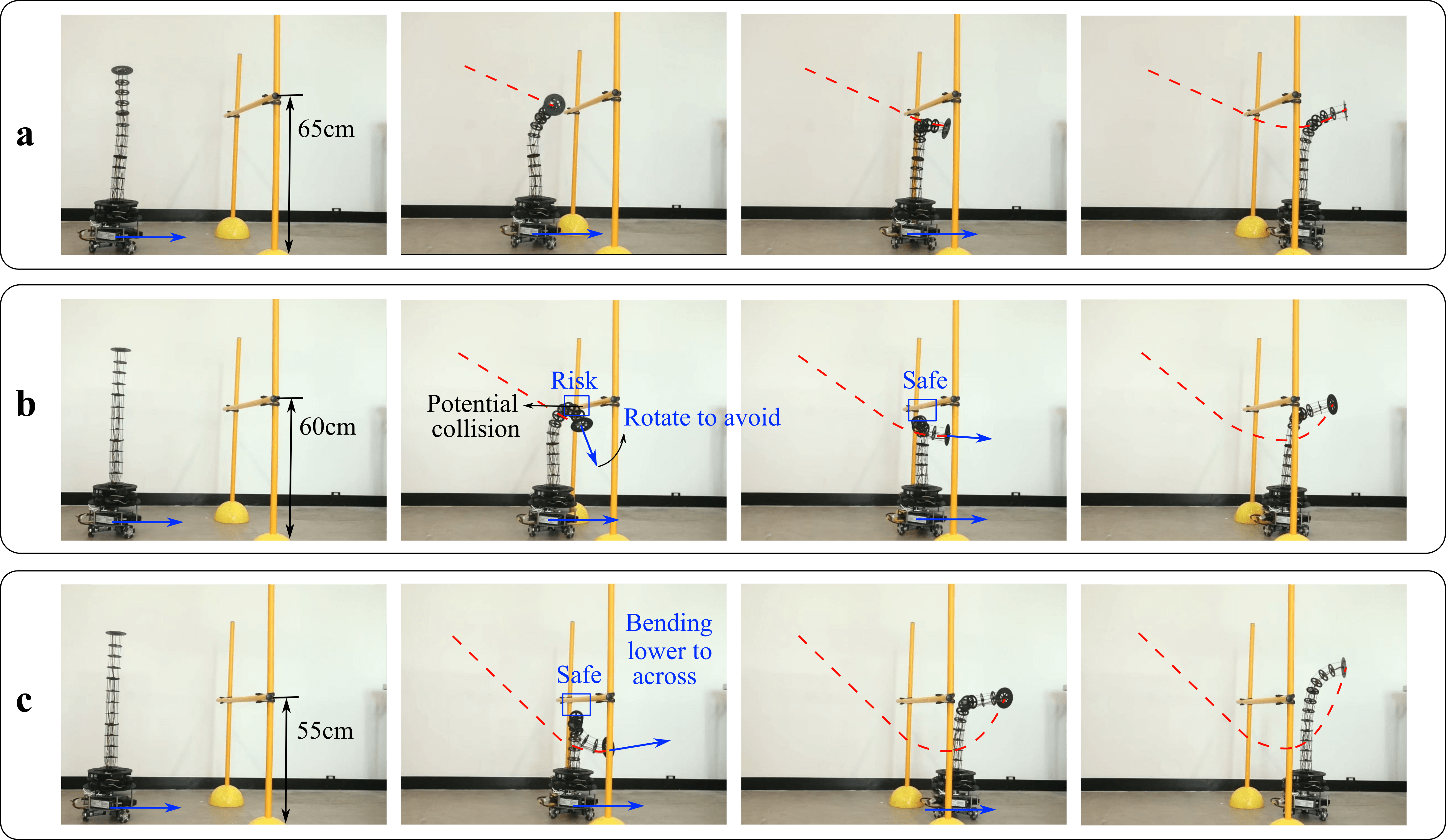}
    \caption{Experiment 4: Hurdle crossing experiment. Snapshots of the continuum robot navigating over obstacles at three different heights: (a) 65 cm, (b) 60 cm, (c) 55 cm.}
    \label{fig:experiment5}
\end{figure*}

\begin{figure*}[t]
    \centering    \includegraphics[width=0.9\linewidth]{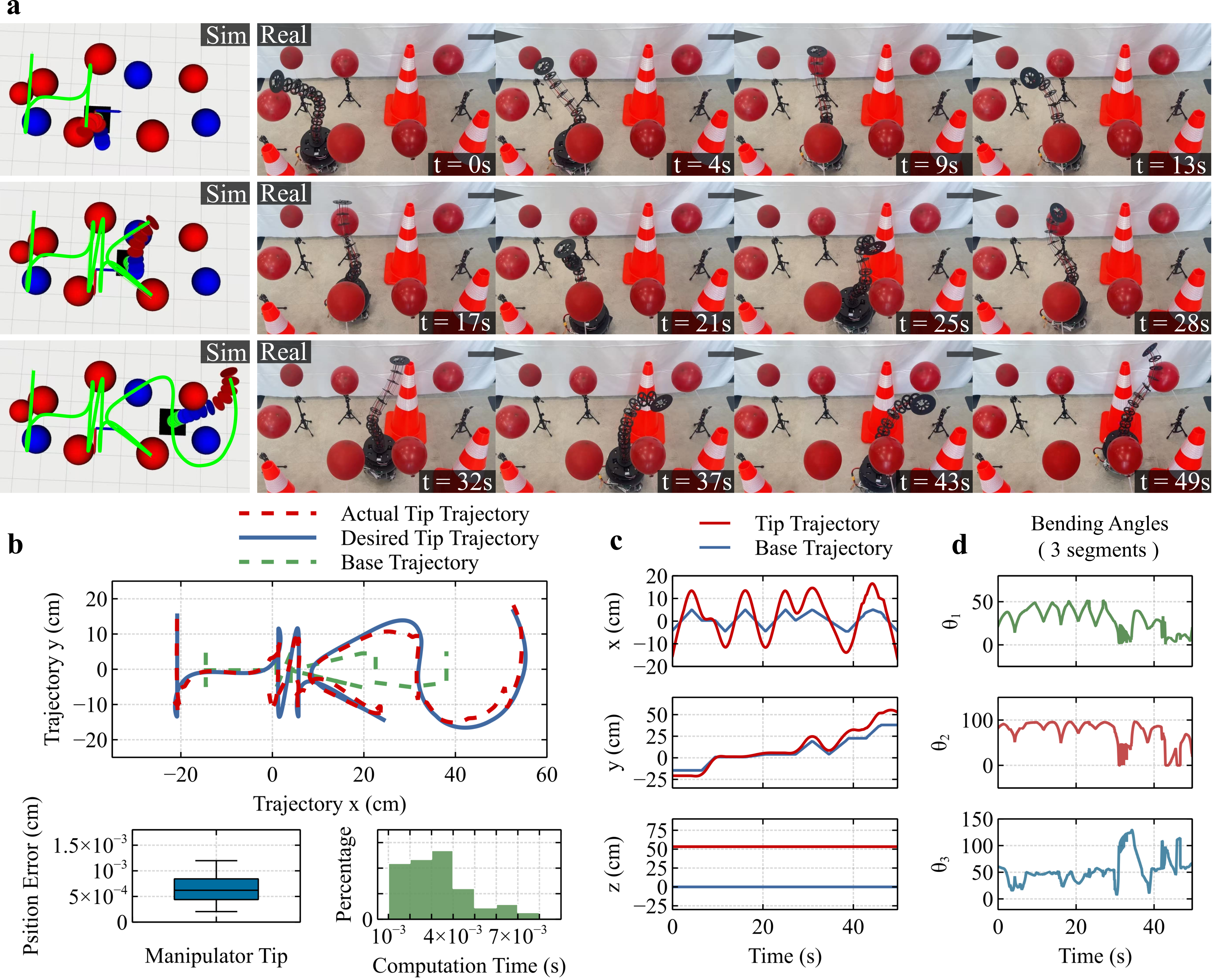}
    \caption{Experiment 5: Collaborative tracking of the letter-shaped trajectory in a cluttered environment. Top: Experimental snapshots and simulated trajectory. Bottom: Planar trajectory view with performance metrics including position errors, bending angles, and computation time.}
    \label{fig:experiment4}
\end{figure*}

\begin{figure*}[t]
    \centering   \includegraphics[width=0.9\linewidth]{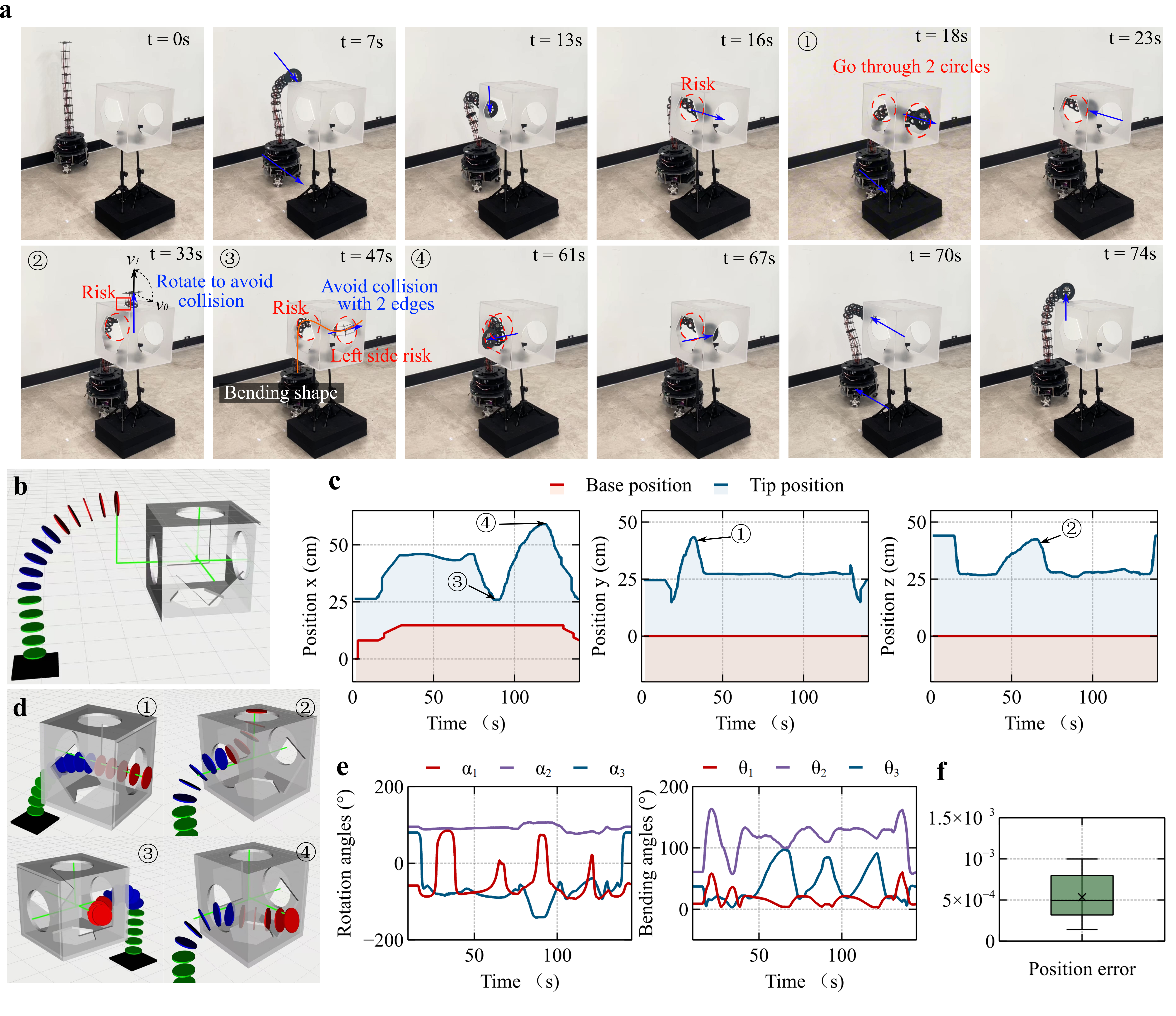}
    \caption{Experiment 6: Box traversal with mobile continuum robot. (a) Motion snapshots, (b) End-effector trajectory inside the box (green), (c) Base and tip position evolution, (d) Four distinct traversal configurations, (e) Configuration angle profiles over time, (f) Position error distribution.}
    \label{fig:experiment6}
\end{figure*}

\subsubsection{Circular trajectory tracking experiment}
The second experiment is designed to evaluate planner performance in environments with increased obstacle density. As illustrated in Fig.~\ref{fig:experiment2}(a), the sequential snapshots clearly demonstrate the manipulator's adaptive behavior: during $t = 5\text{–}15~\mathrm{s}$, it exhibits controlled bending to avoid the smaller obstacle before recovering to its nominal configuration to continue trajectory tracking, while during $t = 20\text{–}30~\mathrm{s}$, it successfully avoids the larger obstacle through similar shape adaptation, ultimately reaching the trajectory endpoint while maintaining continuous motion.

Fig.~\ref{fig:experiment2}(b) provides a comprehensive 3D visualization of the intermediate configurations during the motion, particularly highlighting the pronounced bending deformation away from the red obstacles. This reactive bending ensures collision avoidance while simultaneously maintaining the end-effector position on the desired circular trajectory, successfully balancing safety constraints with tracking accuracy in a complex multi-obstacle environment.

The trajectory tracking performance is quantitatively shown in Fig.~\ref{fig:experiment2}(c), which compares the desired circular path with the estimated end-effector positions from the physical manipulator. The close alignment between these trajectories demonstrates the capability to accurately follow complex geometric paths, even in the presence of avoidance maneuvers. This is further supported in Fig.~\ref{fig:experiment2}(d), where the upper three plots show the evolution of Cartesian positions $p_x$, $p_y$, and $p_z$ over time, revealing consistent tracking of the desired sinusoidal patterns characteristic of circular motion. The lower two plots present the IMU-measured pitch and roll angles of the three segment endpoints, confirming maintained motion smoothness throughout the complex task, with continuous orientation variations and absence of abrupt changes during the avoidance maneuvers.

This experiment extends the validation scope by introducing increased geometric complexity through circular trajectory tracking and multiple obstacles of varying sizes. Experiment 1 demonstrates basic interactive control and single-obstacle avoidance in simple reciprocating motion, and the current scenario validates that the system can handle sustained shape adaptation, sequential obstacle avoidance, and complex multi-axis coordination.

\subsubsection{Trajectory Tracking through Barrier Plates}

Fig.~\ref{fig:experiment3}(b) illustrates an extended task in a more challenging setting, where the manipulator sequentially traverses a red circular-hole barrier plate and a blue square-hole barrier plate, and then performs a helical trajectory. The snapshots capture the manipulator’s continuous adaptation as it navigates through these spatial constraints, exhibiting coordinated shape deformation and smooth motion continuity throughout the process. This experiment further illustrates the capability of the system to maintain precise end-effector control even when multiple narrow passages restrict the available configuration space.

% \begin{figure}[t]
%     \centering   \includegraphics[width=\linewidth]{images/hurdle.png}
%     \caption{Hurdle crossing performance with \SI{60}{\centi\meter} height: Trajectory variation, end-effector orientation, and bending angle profiles during obstacle navigation.}
%     \label{fig:experiment5-data}
% \end{figure}

The quantitative evaluation of three representative trajectories—a four-leaf clover, a pentagram, and a helix—is summarized in the subplots of Fig.~\ref{fig:experiment3}. The four-leaf clover and pentagram trajectories are executed using a single barrier plate (Fig.~\ref{fig:experiment3}(a)), whereas the helical trajectory is performed in a more complex setup involving two barrier plates (Fig.~\ref{fig:experiment3}(b)). The results confirm accurate reproduction of complex planar geometries in the $x$--$y$ plane, together with high-fidelity tracking of positions $\mathbf{p} = (p_x, p_y, p_z)$, stable real-time posture adjustment $\mathbf{o} = (o_x, o_y, o_z)$ for collision avoidance, and responsive bending control $\boldsymbol{\alpha} = (\alpha_1, \alpha_2, \alpha_3)$, which together ensure smooth shape transitions and stable end-effector behavior.

These experiments collectively demonstrate the effective coordination between high-level planning and low-level deformation control in regulating both shape and orientation during motion. This coordination enables the overall framework to achieve precise and collision-free trajectory execution in confined environments with complex geometric constraints.

\subsection{Coordinated Motion with Mobile Base in Cluttered Environments}
\label{subsec:mobile_base}
While the fixed-base experiments validate the framework’s fundamental obstacle avoidance and adaptive control capabilities, practical deployment often requires broader workspace coverage and dynamic repositioning. To address this, the proposed framework is integrated with a mobile continuum robot mounted on an omnidirectional base, enabling coordinated base–arm motion for enhanced dexterity and reachability.

This mobile-base configuration allows the manipulator not only to deform for local obstacle avoidance but also to translate its base to achieve global path optimization in complex, cluttered environments. Such coordination is particularly valuable in scenarios where fixed-base manipulators are limited by workspace boundaries.

\subsubsection{Hurdle crossing experiment}
To evaluate the adaptability and coordination of the proposed framework, Experiment~4 tested the mobile continuum robot traversing a horizontal hurdle set at three decreasing heights: \SI{65}{\centi\meter}, \SI{60}{\centi\meter}, and \SI{55}{\centi\meter}. The robot, mounted on an omnidirectional base, moved from one side to the other, requiring whole-body coordination to avoid collisions. As the base advanced, the arm detects imminent contact and adaptively bent downward, allowing safe traversal while the base adjusted its position to maintain feasible arm configurations. Snapshots in Fig.~\ref{fig:experiment5} show successful traversal for different hurdle heights.

\subsubsection{Collaborative tracking experiment in cluttered environments}
As illustrated in Fig.~\ref{fig:experiment4}, the experiment involves tracking a continuous trajectory that smoothly connects three letters in a cluttered environment with densely distributed obstacles. The top panel shows snapshots of the physical experiment, together with a simulated visualization of the complete letter-shaped trajectory. During the task, the continuum robot and the mobile base coordinate their motions, with the base repositioning smoothly to maintain feasible arm configurations and avoid collisions with ground-level obstacles.

Obstacle avoidance is achieved through real-time shape adaptation of the manipulator, enabling continuous trajectory execution despite frequent close-proximity interactions with surrounding obstacles. The manipulator successfully completes the entire trajectory without any collisions, demonstrating robust coordination between motion planning and deformation control in a highly constrained environment.

The bottom panels of Fig.~\ref{fig:experiment4} provide quantitative results. The $x$–$y$ trajectories show close agreement between the desired and actual end-effector paths, together with coordinated base movement. The position profiles $(p_x, p_y, p_z)$ further confirm smooth and synchronized base–arm motion. The corresponding bending angles $(\theta_1, \theta_2, \theta_3)$ illustrate continuous shape regulation in response to obstacle proximity.

The computation time distribution shows that most control cycles are completed within $5.0 \times 10^{-3}$ \si{\second}, corresponding to an effective control frequency above \SI{200}{\hertz}. These results verify the feasibility and robustness of the proposed coordinated motion framework for accurate and collision-free operation in cluttered environments.

\begin{figure*}[!htbp]
\centering
\includegraphics[width=0.8\linewidth]{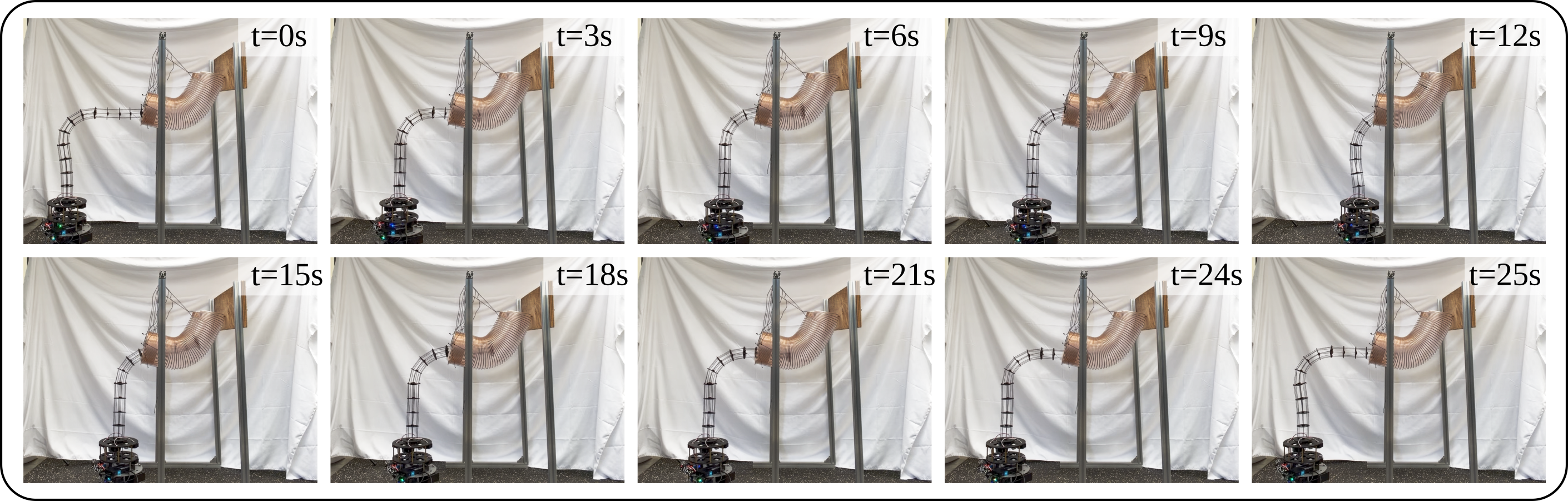}  % ← 换成你的实际图片路径
\caption{Experiment 7: Pipe traversal with the mobile continuum robot. Sequential snapshots demonstrate coordinated base–arm motion and continuous shape adaptation, enabling smooth and collision-free passage through the constrained pipeline.}
\label{fig:pipe_traversal}
\end{figure*}

\subsubsection{Box traversal experiment}
Experiment~6 examined traversal through a confined box structure with multiple internal passages (Fig.~\ref{fig:experiment6}). The robot successfully reconfigured its shape to pass through varying directional openings while maintaining coordinated base–arm motion. Four representative configurations in Fig.~\ref{fig:experiment6}(d) correspond to execution snapshots at critical times (t=18s, 33s, 47s, 61s), validating accurate whole-body planning. Smooth end-effector trajectories (Fig.~\ref{fig:experiment6}(b)), minimal position errors (Fig.~\ref{fig:experiment6}(f)), and coordinated base-tip motion (Fig.~\ref{fig:experiment6}(c)) further demonstrate effective real-time adaptation. Overall, these experiments confirm the capability of the framework to execute complex, constrained navigation tasks with robust planning and control integration.

\subsubsection{Pipe Traversal in Constrained Tubular Environments}
To further demonstrate the framework’s capability in highly constrained, tubular workspaces, we conducted a real-world pipe-traversal experiment with the mobile continuum robot. As shown in Fig.~\ref{fig:pipe_traversal}, the omnidirectional base advances continuously while the continuum arm adaptively bends and reconfigures its shape to maintain collision-free passage through the curved and narrow pipeline. The sequence of snapshots illustrates smooth whole-body coordination: the base provides global progress, while the arm performs real-time local deformation for obstacle avoidance and clearance maintenance.
Despite the limited free space inside the pipe, the robot completes the traversal without interruption or collision, relying solely on the proposed planning and obstacle-avoidance strategy.

\section{Conclusion}
\label{sec:conclusion}
This paper presented a hierarchical motion planning and control framework for mobile-base continuum robots operating in cluttered environments. The architecture integrates global trajectory generation, local motion refinement, and inverse dynamics-based execution to achieve coherent whole-body coordination. The framework is supported by a geometry-consistent FABRIK-based IK solver that preserves segment-level curvature continuity and efficient computational foundation, a Riemannian manifold optimization strategy for feasible orientation planning, and a constraint-aware formulation that embeds safety and kinematic feasibility into both planning and control. Extensive simulation and hardware evaluations across pipe traversal, obstacle-dense settings, and confined-space navigation demonstrate improved tracking performance, computational efficiency, and operational robustness compared with representative baseline approaches.

Future research will extend the framework to dynamic environments with moving obstacles, introduce online adaptation for model uncertainties and physical disturbances, and investigate more scalable optimization strategies for long-horizon tasks. Incorporating multimodal perception, such as tactile or vision-based feedback, also presents a promising
direction for enabling contact-rich manipulation and enhanced environmental awareness. Overall, this work provides a step toward more reliable and autonomous deployment of mobile continuum robots in complex, unstructured environments.

\section{Acknowledgement}
\label{sec:acknow}
This funding is supported by General Research Fund under Grant 17204222.

\bibliographystyle{IEEEtran}
\bibliography{root.bib}

\end{document}